\documentclass[letterpaper]{article} 
\usepackage{aaai2026}  
\usepackage{times}  
\usepackage{helvet}  
\usepackage{courier}  
\usepackage[hyphens]{url}  
\usepackage{graphicx} 
\usepackage{natbib}  
\usepackage{caption} 
\usepackage{algorithm}
\usepackage{algorithmic}

\usepackage{amsmath}
\usepackage{amssymb}
\usepackage{rotating}
\usepackage{multirow}
\usepackage{bigstrut}
\usepackage{booktabs}
\usepackage{supertabular}
\usepackage{newfloat}
\usepackage{listings}
\DeclareCaptionStyle{ruled}{labelfont=normalfont,labelsep=colon,strut=off} 
\floatstyle{ruled}
\newfloat{listing}{tb}{lst}{}
\floatname{listing}{Listing}
\usepackage{hyperref}

\vspace{-10mm}
\title{Leveraging Large-Scale Pretrained Spatial-Spectral Priors for General \\ Zero-Shot Pansharpening}
\author{
Yongchuan Cui\textsuperscript{\rm 1,2},
Peng Liu\textsuperscript{\rm 1,2,}\thanks{Corresponding author.}, 
Yi Zeng\textsuperscript{\rm 3}
}
\affiliations{
    \textsuperscript{\rm 1}Aerospace Information Research Institute, Chinese Academy of Sciences\\
    \textsuperscript{\rm 2}School of Electronic, Electrical and Communication Engineering, University of Chinese Academy of Sciences\\
    \textsuperscript{\rm 3}College of Information, Beijing Forestry University\\
    yongchuancui@gmail.com, liupeng202303@aircas.ac.cn, zengyi@bjfu.edu.cn
}

\usepackage{bibentry}

\begin{document}

\maketitle

\begin{abstract}
Existing deep learning methods for remote sensing image fusion often suffer from poor generalization when applied to unseen datasets due to the limited availability of real training data and the domain gap between different satellite sensors. To address this challenge, we explore the potential of foundation models by proposing a novel pretraining strategy that leverages large-scale simulated datasets to learn robust spatial-spectral priors. Specifically, our approach first constructs diverse simulated datasets by applying various degradation operations (blur, noise, downsampling) and augmentations (bands generation, channel shuffling, high-pass filtering, color jittering, etc.) to natural images from ImageNet and remote sensing images from SkyScript. We then pretrain fusion models on these simulated data to learn generalizable spatial-spectral representations. The pretrained models are subsequently evaluated on six datasets (WorldView-2/3/4, IKONOS, QuickBird, GaoFen-2) using zero-shot and one-shot paradigms, with both full- and freeze-tuning approaches for fine-tuning. Extensive experiments on different network architectures including convolutional neural networks, Transformer, and Mamba demonstrate that our pretraining strategy significantly improves generalization performance across different satellite sensors and imaging conditions for various fusion models. The pretrained models achieve superior results in zero-shot scenarios and show remarkable adaptation capability with minimal real data in one-shot settings. Our work provides a practical solution for cross-domain pansharpening, establishes a new benchmark for generalization in remote sensing image fusion tasks, and paves the way for leveraging foundation models through advanced training strategies.
\end{abstract}

\section{Introduction}

\begin{figure}[t]
  \centering
  \includegraphics[width=1.0\linewidth]{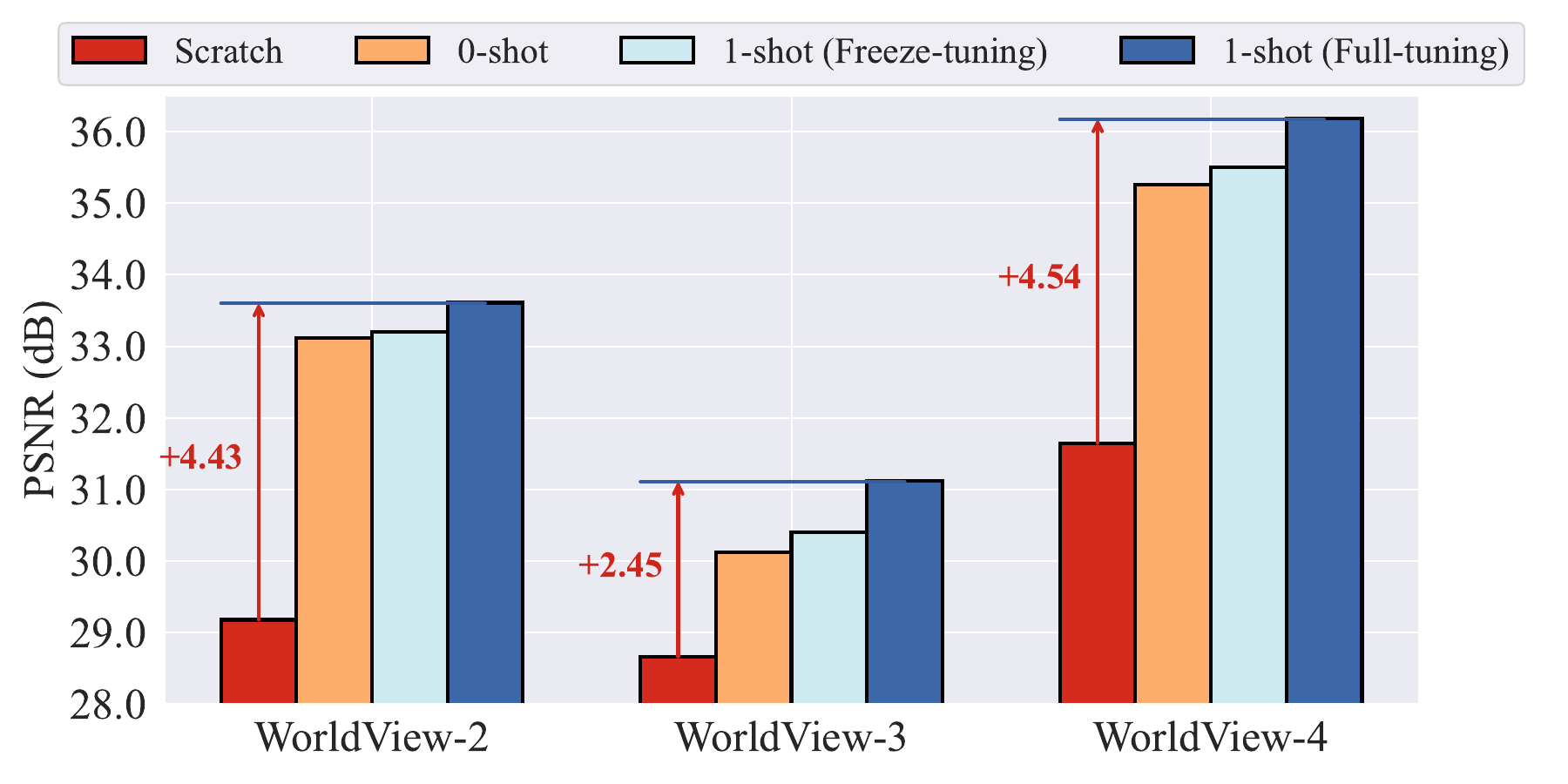}
  \caption{Comparison of PEMAE~\cite{PEMAE} model w/ and w/o pretraining on SkyScript~\cite{SkyScript}.}
  \label{fig:teaser}
\end{figure}

Remote sensing image fusion plays a crucial role in Earth observation by combining complementary information from multiple sources to create comprehensive and high-quality imagery. Among various fusion techniques, pansharpening has emerged as a fundamental task that aims to enhance the spatial resolution of multispectral images (MS) by integrating high-resolution panchromatic imagery (PAN) while preserving spectral characteristics.

The evolution from traditional fusion methods to deep learning-based approaches has marked a significant advancement in the field~\cite{surv1,surv3}. While classical techniques such as component substitution and multiresolution analysis have provided reasonable results, they often struggle with spectral distortion and spatial artifacts~\cite{surv1,surv2}. In recent years, deep learning methods have demonstrated remarkable improvements in both spatial and spectral fidelity, achieving superior performance compared to traditional approaches through their ability to learn complex spatial-spectral relationships~\cite{surv3,surv4}.

Despite the substantial progress made by deep learning models, they suffer from significant generalization challenges when applied to unseen datasets or different satellite sensors. The domain gap between training and testing scenarios, limited availability of real training data, and variations in imaging conditions across different sensors pose substantial obstacles to robust performance~\cite{ZeroSharpen,ZSL,Xing_2025_CVPR,sun2025decadedeeplearningremote}. Large-scale foundation models have shown exceptional generalization capabilities in various computer vision tasks through their massive parameter sizes and extensive training data, effectively spanning vast sample spaces. Nevertheless, in the pansharpening domain, the lack of large-scale training datasets has hindered the development and adoption of foundation models, preventing the field from benefiting from their superior generalization properties. The most convincing evidence is that there still hasn't emerged any large-scale foundation model with massive parameters specifically designed for pansharpening tasks.

Therefore, rather than focusing on model architecture design, this work emphasizes dataset synthesis and pretraining strategies for pansharpening models. Our primary objective is to explore large-scale pretraining approaches specifically tailored for pansharpening tasks. We propose learning spatial-spectral priors from large-scale simulated datasets and conduct extensive experiments across diverse network architectures including convolutional neural networks (CNNs)~\cite{alexnet}, Transformers~\cite{ViT}, and Mamba~\cite{mamba}, demonstrating that large-scale simulated pretraining significantly enhances model generalization capabilities. To rigorously evaluate the model's generalization and adaptation to real datasets, we conducted challenging zero-shot (pretraining only) and one-shot (fine-tuning on one single image) experiments. As shown in Fig.~\ref{fig:teaser}, the results on WorldView-2/3/4~\cite{surv2} datasets demonstrate that pretraining achieves significantly better performance compared to training from scratch, and full fine-tuning (updating all parameters) outperforms freeze fine-tuning (only updating the final convolutional layer). The results show promising performance in both zero-shot and one-shot learning scenarios, establishing a practical foundation for cross-domain pansharpening applications and the development of foundation models for pansharpening. To the best of our knowledge, this is the first work to explore the potential of large-scale pretraining for pansharpening tasks.

\section{Related Work and Motivation}

\subsection{Deep Learning Methods for Pan-sharpening}

Traditional pansharpening methods, including component substitution~\cite{GS,PCA1,IHS1}, multi-resolution analysis~\cite{MTF-GLP1,MTF-GLP2}, and variational optimization~\cite{FE-HPM,VO+Net} approaches, have been extensively studied but face limitations in spectral-spatial fidelity and computational efficiency~\cite{surv1,surv2}. The landscape of pansharpening underwent a paradigm shift with the introduction of deep learning methodologies, fundamentally transforming the approach from handcrafted feature engineering to data-driven representation learning~\cite{surv3,surv4,10696963}.

Pioneering convolutional architectures demonstrated unprecedented capabilities in capturing complex spatial-spectral correlations. Models like PNN~\cite{PNN}, PanNet~\cite{PanNet} and MSDCNN~\cite{MSDCNN} established the foundation for learning-based fusion strategies, proving that neural networks could effectively approximate the intricate spectral mapping by hierarchical feature extraction. Since then, various deep learning models spring up. For instance, the introduction of Transformer architectures~\cite{transformer,ViT} marked a significant milestone, offering the capability to model long-range dependencies through self-attention mechanisms. Specialized Transformer variants for pansharpening~\cite{meng2022vision,BDT,PEMAE,Hypertransformer,10972611} have demonstrated remarkable performance improvements. To address the quadratic computational complexity of Transformers, various Mamba~\cite{mamba,mamba2} models that combine structured state space sequence models with dynamic selective mechanisms for more efficient computation have emerged, such as PanMamba~\cite{PanMamba}, FusionMamba~\cite{FusionMamba} and S$^2$CMamba~\cite{S2CMamba}. Generative modeling approaches also emerged as a promising direction for enhancing the fidelity of fused outputs. Generative adversarial networks~\cite{cui2024reconstruction,Pan-GAN,PSGAN,Mun-GAN} and diffusion-based methods~\cite{DDIF,SSDiff,PanDiff,rui2024unsupervised,xing2024empower,xing2024crossdiff} have shown particular promise in generating more realistic and detailed fused imagery.

\subsection{Motivation}

Despite the efforts, the generalization of deep learning-based pansharpening models in real-world scenarios remains challenging. Methods such as unsupervised learning~\cite{Pan-GAN,Lambda-PNN,rui2024unsupervised,cui2025overcomingidentitymappingproblem} paradigms to remove the need for paired data, generalized module design~\cite{xing2024empower,Xing_2025_CVPR,Cui_2025_ICCV} strategies to enhance robustness have been proposed to address this issue. However, these approaches often require extensive architectural modifications or computationally intensive retraining, which limits their practicality for real-world deployment.

The limitations of current approaches motivate our exploration of large-scale pretraining strategies for pansharpening. While foundation models have demonstrated exceptional generalization capabilities in various computer vision domains through massive parameter spaces and extensive training data, the pansharpening field has not fully leveraged these advantages due to the scarcity of large-scale training datasets. This gap presents an opportunity to develop novel pretraining approaches and appropriate datasets construction methods that can bridge the domain-specific challenges of pansharpening with the generalization benefits of foundation models.

Motivated by the observation that a large volume of simulated data can provide a rich and diverse training environment for learning robust spatial-spectral representations, our work is the first to explore simulated pretraining for learning spectral-spatial priors in pansharpening. By constructing large-scale simulated datasets through various degradation operations and augmentations, we create a training space that captures cross-sensor variations and geographic diversity while providing sufficient data volume for effective pretraining. This approach enables us to explore the potential of foundation models in pansharpening without requiring massive amounts of real training data, and paves the way for developing more generalizable pansharpening models.

\section{Proposed Method}

To address the scarcity of real pansharpening training data and enhance the generalization ability of deep models, we construct simulated datasets by applying various degradation operations and augmentations to large-scale images.


To investigate the differences between natural image and remote sensing image pretraining, we employ two distinct data sources: natural images from ImageNet~\cite{imagenet} and remote sensing images from SkyScript~\cite{SkyScript}. Natural images focus more on spatial details and rich color information, providing high-resolution features, rich texture patterns, diverse semantic content, and extensive visual structures that help models learn generalizable spatial-spectral representations, while remote sensing images from SkyScript~\cite{SkyScript}, a multi-source dataset encompassing diverse satellite sensors including high-resolution commercial satellites (Planet SkySat), multispectral satellites (Landsat 8/9, Sentinel-2), and aerial imagery from various geographical regions, offer domain-specific characteristics, authentic spectral-spatial relationships, and real-world radiometric data that are more closely aligned with real pansharpening scenarios. 

\subsection{Simulated Dataset Construction}

Due to the non-uniformity of data channels across different sources, such as ImageNet~\cite{imagenet} containing only RGB channels and products from different satellites carrying various sensors with different numbers of spectral bands, we first simulate the generation of additional spectral bands, setting the maximum number of bands to $C_{\text{max}}$.

\subsubsection{Multispectral Image Synthesis}

For a given image $\mathbf{I} \in \mathbb{R}^{C \times H \times W}$ with $C$ channels where $H$ and $W$ represent the height and width of the image in pixels, respectively (e.g., RGB images with $C = 3$ or Landsat-7 ETM+ with $C = 7$ excluding panchromatic band), we generate a simulated higher-dimensional multispectral image $\mathbf{I}_{\text{MS}} \in \mathbb{R}^{C_{\text{max}} \times H \times W}$ by synthesizing additional spectral bands through linear combinations of existing channels:

\begin{equation}
    \mathbf{I}_{\text{MS}}^{(C+1:C_{\text{max}})} = \mathbf{A}\mathbf{I},
\end{equation}
where $\mathbf{A} = [\alpha_{ij}] \in \mathbb{R}^{(C_{\text{max}}-C) \times C}$ is a randomly generated weight matrix with each row normalized to sum to unity:

\begin{equation}\label{eq:ms_syn}
\alpha_{ij} = \frac{\tilde{\alpha}_{ij}}{\sum_{k=1}^{C} \tilde{\alpha}_{ik}}, \quad \tilde{\alpha}_{ij} \sim \text{Uniform}(0, 1).
\end{equation}

For a specific band in the synthesized multispectral image, we have:
\begin{equation}
\mathbf{I}_{\text{MS}}^{(i)} = \sum_{j=1}^{C} \alpha_{ij} \mathbf{I}^{(j)}, \quad i = C+1, C+2, \ldots, C_{\text{max}},
\end{equation}
where $\mathbf{I}^{(j)}$ represents the $j$-th channel of the original image and $\mathbf{I}_{\text{MS}}^{(i)}$ represents the $i$-th channel of the synthesized multispectral image.

\subsubsection{Simulated PAN Image Generation}

Given a multispectral image $\mathbf{I}_{\text{MS}} \in \mathbb{R}^{C \times H \times W}$ with $C$ spectral bands, similar to the multispectral image synthesis, we simulate a panchromatic image $\mathbf{I}_{\text{PAN}} \in \mathbb{R}^{1 \times H \times W}$ through a weighted linear combination of a random subset of spectral bands:

\begin{equation}
\mathbf{I}_{\text{PAN}} = \sum_{i \in \mathcal{S}} w_i \mathbf{I}_{\text{MS}}^{(i)},
\end{equation}
where $\mathcal{S} \subset \{1, 2, \ldots, C\}$ is a randomly selected subset of spectral bands with $|\mathcal{S}| \in [2, C]$, $|\cdot|$ denotes the cardinality of a set, and $\mathbf{w} = [w_1, w_2, \ldots, w_{|\mathcal{S}|}]$ are randomly generated weights that sum to unity:

\begin{equation}\label{eq:pan_syn}
w_i = \frac{\tilde{w}_i}{\sum_{j \in \mathcal{S}} \tilde{w}_j}, \quad \tilde{w}_i \sim \text{Uniform}(0, 1).
\end{equation}

Eq. (\ref{eq:pan_syn}) mimics the physical process where panchromatic sensors capture a weighted combination of multiple spectral bands, with the weights varying depending on the sensor's randomly generated spectral response characteristics.

\subsection{Training Strategy}

Contemporary supervised pansharpening methods predominantly adopt Wald's protocol~\cite{wald} for degradation-reconstruction proxy learning, where multispectral and panchromatic images undergo fixed-ratio downsampling and blurring (simulating modulation transfer function and point spread function characteristics) to establish low-to-high spectral mapping, with original MS as ground truth (GT) signal. While effective for learning spectral-spatial correspondences, this rigid degradation pipeline severely limits generalization capability. Networks risk memorizing sensor-specific deblurring/upsampling patterns rather than learning transferable restoration principles. To address this, we design various stochastic degradation and augmentation operations. This forces the network to learn generalized representations of cross-resolution spectral fidelity and spatial detail injection, rather than overfitting to deterministic degradation patterns.

Given a clean multispectral image $\mathbf{I}_{\text{MS}} \in \mathbb{R}^{C_{\text{max}} \times H \times W}$ and a clean panchromatic image $\mathbf{I}_{\text{PAN}} \in \mathbb{R}^{1 \times H \times W}$, we generate degraded inputs via a composite transformation:
\begin{equation}
(\mathbf{I}_{\text{MS}}^{\text{deg}}, \mathbf{I}_{\text{PAN}}^{\text{deg}}) = \mathcal{D} \circ \mathcal{A}(\mathbf{I}_{\text{MS}}, \mathbf{I}_{\text{PAN}}),
\end{equation}
where $\mathcal{A}$ denotes the augmentation operator, $\mathcal{D}$ denotes the degradation operator, and $\circ$ represents function composition.

\subsubsection{Degradation Operator $\mathcal{D}$}

The degradation operator is defined as a composition:
\begin{equation}
\mathcal{D} = \mathcal{B}_\xi \circ \mathcal{D}_s \circ \mathcal{N}_\psi,
\end{equation}
where $\mathcal{B}_\xi$ is a blur operator with kernel $\xi$,
$\mathcal{D}_s$ is a downsampling operator with scale factor $s$, and $\mathcal{N}_\psi$ is a noise operator parameterized by type $\psi$.

\paragraph{(1) Blur Operator $\mathcal{B}_\xi$:}
\begin{equation}
\mathbf{I}^{\text{blur}} = \mathcal{B}_\xi(\mathbf{I}) = \mathcal{K}_\xi * \mathbf{I},
\end{equation}
where $*$ denotes convolution, $\mathcal{K}_\xi$ is randomly chosen from:
\begin{equation}
\mathcal{K}_\xi \in \{\mathcal{G}_{\sigma,k}, \mathcal{B}_k, \mathcal{M}_{\theta,d,k}, \mathcal{M}_k\},
\end{equation}
with Gaussian $\mathcal{G}_{\sigma,k}$, Box $\mathcal{B}_k$, Motion $\mathcal{M}_{\theta,d,k}$, and Median $\mathcal{M}_k$ blur kernels, and uniformly sampled parameters:
\[
\sigma \in [0.1,3.0], \quad \theta \in [0^\circ, 360^\circ], \quad d \in [-1,1], \quad k \in \{3,5,7\}.
\]

\paragraph{(2) Downsampling Operator $\mathcal{D}_s$:}
\begin{equation}
\mathbf{I}^{\text{down}} = \mathcal{D}_s(\mathbf{I}), \quad s \in [0.1, 0.5],
\end{equation}
where $s$ represents the down scale factor (we set $2\times$ to $10\times$ downsampling to simulate multi-scale degradation scenarios), and $\mathcal{D}_s$ uses a random interpolation method from:
\[
\{\text{Nearest}, \text{Bilinear}, \text{Bicubic}, \text{Area}\}.
\]

\paragraph{(3) Noise Operator $\mathcal{N}_\psi$:}
\begin{equation}
\mathbf{I}^{\text{noisy}} = \mathbf{I} + \mathcal{N}_\psi(\mathbf{I}),
\end{equation}
with $\mathcal{N}_\psi \in \{\mathcal{N}_{\text{G}}, \mathcal{N}_{\text{SP}}, \mathcal{N}_{\text{P}}, \mathcal{N}_{\text{S}}\}$ is a random noise operator:
\begin{align*}
\mathcal{N}_{\text{G}}(\mathbf{I}) &= \mathcal{N}(0, \sigma^2), && \sigma \in [0.01, 0.1] \quad \text{(Gaussian)} \\
\mathcal{N}_{\text{SP}}(\mathbf{I}) &= \mathcal{S}(p), && p \in [0.001, 0.01] \quad \text{(Salt \& Pepper)} \\
\mathcal{N}_{\text{P}}(\mathbf{I}) &= \frac{\mathcal{P}(\lambda \mathbf{I})}{\lambda}, && \lambda \in [10, 50] \quad \text{(Poisson)} \\
\mathcal{N}_{\text{S}}(\mathbf{I}) &= \mathbf{I} \odot \mathcal{N}(1, \sigma^2), && \sigma \in [0.05, 0.2] \quad \text{(Speckle)}
\end{align*}
where $\odot$ denotes element-wise multiplication.

\subsubsection{Augmentation Operator $\mathcal{A}$}

The augmentation operator $\mathcal{A}$ is composed of:
\begin{equation}
\mathcal{A} = \mathcal{A}_{\text{spatial}} \circ \mathcal{A}_{\text{spectral}},
\end{equation}
where $\mathcal{A}_{\text{spatial}}$ includes geometric transformations, and $\mathcal{A}_{\text{spectral}}$ applies spectral perturbations.

\paragraph{(1) Spatial Augmentations $\mathcal{A}_{\text{spatial}}$:}
\begin{equation}
\mathcal{A}_{\text{spatial}} \in \{\text{Flip}_{x,y}, \text{Rotate}_{k \cdot 90^\circ}\}, \quad k \in \{0,1,2,3\},
\end{equation}
for horizontal and vertical flips, and discrete rotations.

\paragraph{(2) Spectral Augmentations $\mathcal{A}_{\text{spectral}}$} includes:
\begin{itemize}
    \item \textit{Channel Shuffle:}
    \begin{equation}
    \mathbf{I}^{\text{aug}} = \mathbf{I}[\pi], \quad \pi \sim \text{Permutation}(\{1, \ldots, C_{\text{max}}\}).
    \end{equation}

    \item \textit{Channel Masking:}
    \begin{equation}
    \mathbf{I}^{\text{aug}}_i =
    \begin{cases}
    0, & i \in \mathcal{M}, \\
    \mathbf{I}_i, & \text{otherwise},
    \end{cases}
    \quad \mathcal{M} \subset \{1, \ldots, C_{\text{max}}\}.
    \end{equation}

    \item \textit{High-Pass Filtering (PAN only):}
    \begin{equation}
    \mathbf{I}^{\text{aug}} = \mathcal{H}_\zeta * \mathbf{I}, \quad
    \mathcal{H}_\zeta \in \{\mathcal{L}_3, \mathcal{G}_{0.5,3} - \mathcal{G}_{2.0,7}, \mathcal{S}, \mathcal{C}\},
    \end{equation}
    where $\mathcal{L}_3$ is Laplacian filter with kernel size of $3$, and the DoG filter ($\mathcal{G}_{0.5,3} - \mathcal{G}_{2.0,7}$) uses $\sigma=0.5/2.0$ and kernel sizes $3/7$. $\mathcal{S}$ and $\mathcal{C}$ are Sobel and Canny filters.

    \item \textit{Color Jittering:}
    \begin{equation}
    \mathcal{T}_{\text{color}} = \mathcal{T}_{\text{bri}} \circ \mathcal{T}_{\text{con}} \circ \mathcal{T}_{\text{sat}} \circ \mathcal{T}_{\text{hue}},
    \end{equation}
    applied sequentially to simulate radiometric changes, where $\mathcal{T}_{\text{color}}$ represents the color transformation function composed of brightness $\mathcal{T}_{\text{brightness}}$, contrast $\mathcal{T}_{\text{contrast}}$, saturation $\mathcal{T}_{\text{saturation}}$, and hue $\mathcal{T}_{\text{hue}}$ transformations.
\end{itemize}

\subsubsection{Loss Function}

During pretraining and fine-tuning, we employ $\mathcal{L}_1$ loss to measure the pixel-wise difference between the predicted image and the ground truth:

\begin{equation}
\mathcal{L}_{\text{1}} = \frac{1}{C_{\text{max}} \times H \times W} \sum_{c,h,w} |\mathbf{I}_{\text{MS}}^{(c,h,w)} - \mathcal{F}(\mathbf{I}_{\text{MS}}^{\text{deg}}, \mathbf{I}_{\text{PAN}}^{\text{deg}})^{(c,h,w)}|,
\end{equation}
where $\mathbf{I}_{\text{MS}}$ represents the original high-resolution multispectral image (HRMS) as ground truth, $\mathbf{I}_{\text{MS}}^{\text{deg}}$ and $\mathbf{I}_{\text{PAN}}^{\text{deg}}$ denote the degraded low-resolution multispectral (LRMS) and panchromatic images as network inputs, $\mathcal{F}(\cdot,\cdot)$ represents the fusion network that takes both degraded images as input, and $C_{\text{max}}$ is the maximum number of spectral bands. To handle varying numbers of bands, the network will use fixed $C_{\text{max}}$ channels of input and output. Zero-padding is applied to input images in case of bands less than $C_{\text{max}}$.

\subsection{Zero-shot Generalization and One-shot Tuning}

To evaluate the generalization and adaptation capabilities of pretrained pansharpening models, we consider two practical settings: \textit{zero-shot generalization} and \textit{one-shot tuning}.

In the zero-shot setting, the pretrained model $\mathcal{F}_{\text{pre}}$ is directly applied to unseen satellite datasets without any fine-tuning:
\begin{equation}
\hat{\mathbf{I}}_{\text{MS}} = \mathcal{F}_{\text{pre}}(\mathbf{I}_{\text{MS}}^{\text{deg}}, \mathbf{I}_{\text{PAN}}^{\text{deg}}),
\end{equation}
where $\mathbf{I}_{\text{MS}}^{\text{deg}}$ and $\mathbf{I}_{\text{PAN}}^{\text{deg}}$ are the degraded inputs using the widely used Wald's protocol~\cite{wald}  and $\hat{\mathbf{I}}_{\text{MS}}$ is the predicted fused image.

In the one-shot setting, we fine-tune the pretrained model using only a single pair of degraded inputs. Two strategies are considered:

\textit{(1) Full-tuning}: All model parameters are updated to minimize the $\mathcal{L}_1$ loss:
\begin{equation}
\mathcal{L}_{\text{full}} = \mathcal{L}_{1}(\mathcal{F}_{\text{full}}(\mathbf{I}_{\text{MS}}^{\text{deg}}, \mathbf{I}_{\text{PAN}}^{\text{deg}}), \mathbf{I}_{\text{MS}}).
\end{equation}

\textit{(2) Freeze-tuning}: Only the final output layer is updated while keeping all backbone parameters frozen:
\begin{equation}
\mathcal{L}_{\text{freeze}} = \mathcal{L}_{1}(\mathcal{F}_{\text{freeze}}(\mathbf{I}_{\text{MS}}^{\text{deg}}, \mathbf{I}_{\text{PAN}}^{\text{deg}}), \mathbf{I}_{\text{MS}}).
\end{equation}

This lightweight adaptation enables robust cross-domain performance with minimal tuning data. Especially, the challenging one-shot tuning uses only a single pair of inputs.

\section{Experiments}

\subsection{Setups}

In our experiments, we conduct pretraining on several representative pansharpening models with different architectures, including CNN: FusionNet~\cite{FusionNet}, GPPNN~\cite{GPPNN}, PreMix~\cite{PreMix}, Transformer: PEMAE~\cite{PEMAE}, and Mamba: PanMamba~\cite{PanMamba}, using both the ImageNet~\cite{imagenet} and SkyScript~\cite{SkyScript} datasets. For each satellite dataset, we conduct one-shot experiments using only 10 images (the rest are for validating) and report the results with model ckeckpoint of best PSNR on validtion set. The details of datasets, evaluation metrics, and implementations are provided in the Appendix.

\subsection{Results}

\begin{table}[htbp]
  \centering
  \footnotesize{
    \begin{tabular}{clcc}
    \toprule
    \multicolumn{2}{c}{Model / Dataset} & ImageNet & SkyScript\\
    \midrule
    \multirow{5}[2]{*}{4$\times$} & FusionNet & 38.08/0.9682 & 44.71/0.9840\\
      & GPPNN & 38.61/0.9696 & 46.81/0.9858 \\
      & PreMix & 39.22/0.9701 & 46.64/0.9858 \\
      & PEMAE & \textbf{40.95/0.9726} & \textbf{48.91/0.9870} \\
      & PanMamba & 39.24/0.9707 & 46.87/0.9861 \\
    \midrule
    \multirow{5}[2]{*}{8$\times$} & FusionNet & 33.55/0.9539 & 37.84/0.9767 \\
      & GPPNN & 34.56/0.9588 & 41.60/0.9823 \\
      & PreMix & 35.41/0.9612 & 41.38/0.9824 \\
      & PEMAE & \textbf{37.71/0.9666} & \textbf{44.31/0.9850} \\
      & PanMamba & 36.17/0.9645 & 42.39/0.9838 \\
    \bottomrule
    \end{tabular}%
  }
    \caption{Reconstruction performance (PSNR/SSIM) under $4\times$ and $8\times$ spatial resolution degradations.}
  \label{tab:cmp-pretrain-4x8x}%
\end{table}%

\begin{table}[htbp]
  \centering
  \footnotesize{
    \begin{tabular}{lcccc}
    \toprule
    \multirow{2}[4]{*}{Model} & \multicolumn{3}{c}{\#Params} & \multirow{2}[4]{*}{MACs (G)} \\
  \cmidrule(lr){2-4}      & Total (K) & Tunable (K) & Ratio (\%) &  \\
    \midrule
    FusionNet & 78.63 & 2.31 & 2.94 & 5.13 \\
    GPPNN & 238.59 & 10.24 & 4.29 & 15.64 \\
    PreMix & 616.11 & 41.62 & 6.75 & 40.34 \\
    PEMAE & 388.18 & 2.31 & 0.60 & 20.42 \\
    PanMamba & 497.14 & 2.31 & 0.47 & 14.33\\
    \bottomrule
    \end{tabular}%
    }
  \caption{Comparison of model parameter counts (K) and Multiply-Accumulate Operations (MACs, G).}
  \label{tab:cmp-complexity}%
  \end{table}%

\begin{table*}[htbp]
  \centering
  \footnotesize{
    \begin{tabular}{cccccccc}
    \toprule
    \multicolumn{2}{c}{Model} & WorldView-2 & WorldView-3 & WorldView-4 & IKONOS & QuickBird & GaoFen-2 \\
    \midrule
    \multirow{35}[10]{*}{\begin{sideways}Reduced-Resolution : PSNR$\uparrow$ / SAM$^{\times 10^{-2}}$$\downarrow$ / ERGAS$\downarrow$\end{sideways}} & \multirow{7}[2]{*}{\begin{sideways}FusionNet\end{sideways}} & \textsuperscript{1}30.42/11.1/7.14 & 31.01/11.05/5.45 & 34.89/4.09/2.37 & 35.58/6.37/2.62 & 40.89/3.11/1.46 & 38.85/2.82/1.39 \\
      &   & \textsuperscript{2}32.56/9.62/5.83 & 29.80/12.09/6.58 & 35.09/4.11/2.58 & 35.81/5.24/2.49 & 39.11/3.11/1.99 & 34.26/2.96/2.53 \\
      &   & \textsuperscript{3}32.45/9.52/5.83 & 30.07/12.11/6.24 & 34.72/4.33/2.62 & 35.82/5.32/2.44 & 39.21/3.18/1.94 & 34.00/3.15/2.58 \\
      &   & \textsuperscript{4}32.86/9.26/5.56 & 30.09/11.82/6.28 & 35.44/3.91/2.41 & 36.19/4.95/2.39 & 39.95/2.97/1.80 & 35.26/2.84/2.26 \\
      &   & \textsuperscript{5}32.63/9.32/5.72 & 30.27/11.86/6.06 & 35.00/4.19/2.50 & 36.12/5.15/2.36 & 39.80/3.07/1.80 & 34.69/3.06/2.39 \\
      &   & \underline{\textsuperscript{6}33.58}/\underline{8.39}/4.98 & 30.94/\underline{10.64}/5.52 & \underline{35.97}/\underline{3.56}/\underline{2.17} & 37.08/\underline{4.51}/\underline{2.09} & 42.52/\underline{2.54}/1.25 & \underline{39.50}/\underline{2.38}/\underline{1.32} \\
      &   & \textsuperscript{7}33.56/8.44/\underline{4.97} & \underline{31.07}/10.87/\underline{5.44} & 35.96/3.71/2.17 & \underline{37.10}/4.63/2.10 & \underline{42.68}/2.56/\underline{1.23} & 39.18/2.61/1.35 \\
\cmidrule(lr){2-8}      & \multirow{7}[2]{*}{\begin{sideways}GPPNN\end{sideways}} & \textsuperscript{1}30.82/10.2/6.77 & 30.14/11.62/6.04 & 34.96/3.60/2.42 & 34.56/4.99/2.73 & 40.75/2.78/1.53 & \underline{39.56}/\underline{2.31}/\underline{1.32} \\
      &   & \textsuperscript{2}33.13/8.95/5.35 & 29.49/12.46/7.25 & 34.94/4.14/2.71 & 35.85/5.11/2.49 & 39.43/3.00/1.96 & 34.28/2.91/2.57 \\
      &   & \textsuperscript{3}32.94/9.11/5.55 & 29.61/12.50/7.13 & 34.03/4.57/3.04 & 35.87/5.06/2.48 & 39.28/2.94/1.95 & 34.09/2.87/2.58 \\
      &   & \textsuperscript{4}33.54/8.42/5.05 & 30.35/11.28/6.28 & 35.77/3.82/2.32 & 36.40/4.97/2.31 & 41.19/3.05/1.60 & 36.55/3.00/1.97 \\
      &   & \textsuperscript{5}33.44/8.41/5.21 & 30.42/11.14/6.22 & 35.35/3.82/2.47 & 36.72/4.76/2.23 & 41.50/2.77/1.54 & 36.41/2.61/1.98 \\
      &   & \textsuperscript{6}33.78/\underline{8.26}/4.91 & 30.70/\underline{10.84}/5.87 & 36.16/3.50/2.15 & 36.87/4.72/2.17 & 41.56/2.76/1.41 & 39.48/2.42/1.33 \\
      &   & \underline{\textsuperscript{7}33.89}/8.33/\underline{4.86} & \underline{31.01}/10.86/\underline{5.60} & \underline{36.18}/\underline{3.49}/\underline{2.12} & \underline{37.07}/\underline{4.64}/\underline{2.14} & \underline{42.64}/\underline{2.51}/\underline{1.27} & 39.52/2.32/1.33 \\
\cmidrule(lr){2-8}      & \multirow{7}[2]{*}{\begin{sideways}PreMix\end{sideways}} & \textsuperscript{1}30.46/9.19/7.18 & 29.00/14.11/6.75 & 34.31/3.83/2.59 & 34.45/5.39/2.75 & 40.53/2.91/1.55 & 39.46/2.42/1.33 \\
      &   & \textsuperscript{2}33.59/8.17/4.92 & 30.36/10.41/6.10 & 35.43/3.40/2.36 & 36.45/4.57/2.26 & 39.85/2.51/1.75 & 34.48/2.67/2.33 \\
      &   & \textsuperscript{3}32.91/8.72/5.26 & 29.97/11.08/6.42 & 34.87/3.57/2.53 & 36.01/4.94/2.37 & 39.31/2.76/1.82 & 34.13/2.58/2.39 \\
      &   & \textsuperscript{4}33.82/7.88/4.80 & 30.75/10.17/5.78 & 35.67/3.39/2.26 & 36.86/4.33/2.17 & 40.78/\underline{\textbf{2.38}}/1.57 & 35.57/2.70/2.07 \\
      &   & \textsuperscript{5}33.22/8.51/5.14 & 30.50/10.75/5.98 & 35.26/3.60/2.39 & 36.45/4.64/2.26 & 40.27/2.61/1.64 & 35.07/2.59/2.17 \\
      &   & \underline{\textbf{\textsuperscript{6}34.24}}/\underline{\textbf{7.64}}/\underline{\textbf{4.62}} & \underline{\textbf{31.60}}/\underline{\textbf{9.77}}/\underline{\textbf{5.19}} & \underline{36.25}/\underline{\textbf{3.28}}/\underline{\textbf{2.09}} & \underline{\textbf{37.57}}/\underline{4.23}/\underline{1.99} & \underline{42.69}/2.43/\underline{1.22} & \underline{39.70}/\underline{\textbf{2.29}}/\underline{\textbf{1.28}} \\
      &   & \textsuperscript{7}33.90/8.09/4.78 & 31.34/10.46/5.34 & 36.21/3.43/2.10 & 37.10/4.60/2.09 & 42.45/2.65/1.26 & 39.44/2.33/1.34 \\
\cmidrule(lr){2-8}      & \multirow{7}[2]{*}{\begin{sideways}PEMAE\end{sideways}} & \textsuperscript{1}29.18/15.7/10.1 & 28.66/15.87/8.36 & 31.65/7.22/3.80 & 34.84/5.86/2.71 & 40.96/2.98/1.45 & 37.57/3.19/1.59 \\
      &   & \textsuperscript{2}33.13/8.38/5.26 & 30.23/10.90/6.19 & 35.47/3.69/2.45 & 36.42/4.72/2.25 & 39.89/2.77/1.76 & 34.65/2.81/2.35 \\
      &   & \textsuperscript{3}33.11/8.92/5.45 & 30.12/11.57/6.36 & 35.25/3.98/2.56 & 36.23/4.94/2.37 & 39.99/2.86/1.83 & 34.65/2.73/2.41 \\
      &   & \textsuperscript{4}33.25/8.23/5.19 & 30.43/10.69/6.01 & 35.70/3.63/2.35 & 36.68/4.56/2.19 & 40.40/2.71/1.66 & 35.22/2.76/2.20 \\
      &   & \textsuperscript{5}33.20/8.92/5.42 & 30.40/11.39/6.08 & 35.51/3.94/2.43 & 36.60/4.74/2.27 & 40.67/2.77/1.68 & 35.34/2.65/2.23 \\
      &   & \underline{\textsuperscript{6}33.71}/\underline{7.76}/\underline{4.98} & \underline{31.21}/\underline{10.05}/\underline{5.39} & \underline{\textbf{36.31}}/\underline{3.33}/\underline{2.10} & 37.52/\underline{\textbf{4.20}}/\underline{\textbf{1.99}} & 42.95/\underline{2.41}/1.19 & 39.58/\underline{2.33}/1.32 \\
      &   & \textsuperscript{7}33.61/8.27/5.22 & 31.12/10.63/5.52 & 36.18/3.50/2.11 & \underline{37.52}/4.35/2.02 & \underline{\textbf{43.21}}/2.45/\underline{\textbf{1.16}} & \underline{\textbf{39.71}}/2.38/\underline{1.31} \\
\cmidrule(lr){2-8}      & \multirow{7}[2]{*}{\begin{sideways}PanMamba\end{sideways}} & \textsuperscript{1}21.48/36.0/24.1 & 23.88/25.31/17.4 & 21.19/23.7/17.9 & 24.03/15.4/12.8 & 28.72/13.0/6.43 & 22.32/18.0/10.3 \\
      &   & \textsuperscript{2}32.71/9.93/5.66 & 29.71/12.57/6.71 & 34.58/4.46/2.74 & 35.63/5.62/2.50 & 39.33/3.35/1.90 & 33.95/3.49/2.61 \\
      &   & \textsuperscript{3}32.23/10.9/6.36 & 29.53/13.38/6.99 & 34.09/4.83/2.97 & 35.36/5.70/2.58 & 38.99/3.34/2.00 & 33.69/3.48/2.66 \\
      &   & \textsuperscript{4}32.70/9.93/5.67 & 29.80/12.45/6.58 & 34.62/4.46/2.71 & 35.72/5.57/2.47 & 39.54/3.27/1.85 & 34.13/3.45/2.56 \\
      &   & \textsuperscript{5}32.27/10.9/6.14 & 29.69/13.09/6.72 & 34.13/4.79/2.93 & 35.49/5.60/2.53 & 39.28/3.28/1.93 & 33.93/3.46/2.59 \\
      &   & \underline{\textsuperscript{6}33.31}/\underline{8.89}/\underline{5.11} & \underline{30.76}/\underline{11.02}/\underline{5.67} & \underline{35.13}/\underline{4.10}/\underline{2.37} & \underline{36.28}/\underline{5.44}/\underline{2.33} & 41.41/3.05/1.40 & \underline{37.79}/\underline{2.98}/\underline{1.57} \\
      &   & \textsuperscript{7}33.18/9.18/5.33 & 30.57/11.80/5.88 & 34.53/4.73/2.55 & 36.23/5.50/2.34 & \underline{41.71}/\underline{2.99}/\underline{1.38} & 37.77/3.10/1.61 \\
    \midrule
    \multirow{21}[6]{*}{\begin{sideways}Full-Resolution : $D_{\lambda}^{\times 10^{-1}}$$\downarrow$ / $D_{S}^{\times 10^{-1}}$$\downarrow$ / QNR$\uparrow$\end{sideways}} & \multirow{7}[2]{*}{\begin{sideways}FusionNet\end{sideways}} & \textsuperscript{1}0.37/\underline{0.51}/0.914 & 0.16/\underline{0.81}/\underline{0.904} & 0.17/\underline{0.36}/0.947 & 0.30/\underline{0.66}/\underline{0.906} & 0.23/\underline{0.55}/0.924 & 0.19/0.20/0.962 \\
      &   & \textsuperscript{2}0.18/0.59/0.925 & 0.34/1.13/0.857 & 0.29/0.62/0.912 & 0.65/1.27/0.818 & 1.16/2.02/0.712 & 0.64/1.85/0.765 \\
      &   & \textsuperscript{3}0.19/0.61/0.921 & 0.30/1.04/0.869 & 0.31/0.60/0.912 & 0.61/1.24/0.824 & 1.04/1.94/0.728 & 0.64/1.91/0.760 \\
      &   & \textsuperscript{4}0.19/0.58/0.924 & 0.30/1.05/0.868 & 0.22/0.57/0.923 & 0.61/1.23/0.825 & 0.95/1.84/0.745 & 0.42/1.60/0.806 \\
      &   & \textsuperscript{5}0.17/0.58/\underline{0.925} & 0.28/1.03/0.872 & 0.27/0.60/0.915 & 0.59/1.22/0.828 & 0.92/1.84/0.746 & 0.55/1.82/0.776 \\
      &   & \underline{\textsuperscript{6}0.17}/0.64/0.920 & \underline{\textbf{0.14}}/0.86/0.902 & \underline{0.08}/0.39/0.954 & \underline{0.25}/0.86/0.892 & \underline{0.15}/0.62/\underline{0.924} & \underline{0.03}/\underline{0.09}/\underline{0.988} \\
      &   & \textsuperscript{7}0.18/0.64/0.919 & 0.14/0.84/0.903 & 0.09/0.37/\underline{0.954} & 0.30/0.85/0.889 & 0.21/0.61/0.920 & 0.04/0.18/0.978 \\
\cmidrule(lr){2-8}      & \multirow{7}[2]{*}{\begin{sideways}GPPNN\end{sideways}} & \textsuperscript{1}0.37/\underline{0.52}/0.913 & 0.20/\underline{\textbf{0.67}}/\underline{\textbf{0.914}} & 0.09/\underline{\textbf{0.27}}/\underline{0.965} & \underline{\textbf{0.05}}/\underline{\textbf{0.18}}/\underline{\textbf{0.977}} & \underline{\textbf{0.06}}/\underline{\textbf{0.18}}/\underline{\textbf{0.976}} & 0.02/\underline{\textbf{0.02}}/\underline{\textbf{0.997}} \\
      &   & \textsuperscript{2}0.29/0.79/0.895 & 0.69/1.39/0.802 & 0.22/0.51/0.928 & 0.55/1.17/0.835 & 0.96/1.83/0.745 & 0.49/1.62/0.799 \\
      &   & \textsuperscript{3}0.35/0.78/0.890 & 0.71/1.42/0.799 & 0.41/0.55/0.907 & 0.59/1.18/0.832 & 1.04/1.90/0.731 & 0.63/1.81/0.770 \\
      &   & \textsuperscript{4}0.20/0.73/0.909 & 0.37/1.14/0.854 & 0.10/0.41/0.950 & 0.43/1.09/0.854 & 0.53/1.42/0.816 & 0.32/1.10/0.862 \\
      &   & \textsuperscript{5}0.21/0.69/0.912 & 0.33/1.18/0.854 & 0.23/0.48/0.931 & 0.34/0.97/0.873 & 0.46/1.43/0.820 & 0.32/1.41/0.833 \\
      &   & \underline{\textbf{\textsuperscript{6}0.11}}/0.65/\underline{0.925} & 0.19/0.91/0.891 & \underline{0.06}/0.30/0.964 & 0.28/0.94/0.881 & 0.18/0.72/0.912 & 0.06/0.06/0.988 \\
      &   & \textsuperscript{7}0.15/0.68/0.918 & \underline{0.19}/0.98/0.885 & 0.07/0.34/0.959 & 0.32/0.96/0.876 & 0.16/0.89/0.897 & \underline{\textbf{0.01}}/0.03/0.995 \\
\cmidrule(lr){2-8}      & \multirow{7}[2]{*}{\begin{sideways}PreMix\end{sideways}} & \underline{\textsuperscript{1}0.12}/\underline{\textbf{0.12}}/\underline{\textbf{0.976}} & 0.60/\underline{0.92}/0.854 & 0.31/0.37/0.933 & 0.60/\underline{0.72}/0.872 & 0.32/\underline{0.40}/\underline{0.929} & \underline{0.03}/\underline{0.06}/\underline{0.991} \\
      &   & \textsuperscript{2}0.23/0.76/0.903 & 0.42/1.25/0.839 & 0.14/0.45/0.942 & 0.53/1.16/0.839 & 0.78/1.69/0.771 & 0.38/1.48/0.822 \\
      &   & \textsuperscript{3}0.27/0.74/0.901 & 0.45/1.26/0.835 & 0.27/0.59/0.915 & 0.54/1.20/0.834 & 0.92/1.86/0.745 & 0.60/1.92/0.762 \\
      &   & \textsuperscript{4}0.19/0.71/0.911 & 0.33/1.15/0.856 & 0.07/0.35/0.958 & 0.45/1.08/0.854 & 0.56/1.44/0.812 & 0.18/1.15/0.870 \\
      &   & \textsuperscript{5}0.24/0.69/0.909 & 0.35/1.14/0.855 & 0.18/0.51/0.932 & 0.50/1.16/0.841 & 0.72/1.67/0.778 & 0.38/1.68/0.802 \\
      &   & \textsuperscript{6}0.15/0.68/0.918 & \underline{0.22}/1.01/\underline{0.880} & \underline{\textbf{0.04}}/\underline{0.30}/\underline{\textbf{0.966}} & \underline{0.28}/0.94/\underline{0.881} & \underline{0.14}/0.71/0.916 & 0.03/0.16/0.981 \\
      &   & \textsuperscript{7}0.18/0.68/0.916 & 0.25/1.02/0.876 & 0.06/0.37/0.957 & 0.38/1.06/0.861 & 0.19/0.83/0.900 & 0.03/0.10/0.987 \\
    \midrule
    \multicolumn{3}{c}{Continued to next page.} &   &   &   &   &  \\
    \end{tabular}}%
\end{table*}%

\begin{table*}[t]
  \centering
  \footnotesize{
    \begin{tabular}{cccccccc}
      \midrule
      \multicolumn{2}{c}{Model} & WorldView-2 & WorldView-3 & WorldView-4 & IKONOS & QuickBird & GaoFen-2 \\
      \midrule
      \multirow{14}[4]{*}{\begin{sideways}Full-Resolution\end{sideways}} & \multirow{7}[2]{*}{\begin{sideways}PEMAE\end{sideways}} & \textsuperscript{1}1.00/1.24/0.797 & 0.64/1.00/0.846 & 0.89/0.84/0.841 & 0.56/\underline{0.65}/0.883 & 0.69/\underline{0.33}/\underline{0.901} & 0.59/0.24/0.920 \\
        &   & \textsuperscript{2}0.25/0.68/0.910 & 0.39/1.13/0.852 & 0.25/0.59/0.918 & 0.51/1.08/0.849 & 0.98/1.76/0.747 & 0.72/1.86/0.759 \\
        &   & \textsuperscript{3}0.35/0.66/0.901 & 0.52/1.18/0.836 & 0.32/0.61/0.910 & 0.77/1.24/0.811 & 1.19/1.97/0.713 & 0.73/2.02/0.743 \\
        &   & \underline{\textsuperscript{4}0.25}/0.67/0.910 & 0.36/1.11/0.857 & 0.20/0.56/0.925 & 0.49/1.06/0.852 & 0.86/1.70/0.763 & 0.60/1.78/0.775 \\
        &   & \textsuperscript{5}0.37/0.66/0.900 & 0.50/1.15/0.842 & 0.28/0.59/0.915 & 0.72/1.23/0.816 & 1.08/1.91/0.727 & 0.59/1.91/0.763 \\
        &   & \textsuperscript{6}0.28/0.62/\underline{0.913} & \underline{0.28}/\underline{0.90}/\underline{0.885} & \underline{0.15}/\underline{0.33}/\underline{0.952} & \underline{0.29}/0.77/\underline{0.897} & \underline{0.36}/0.87/0.881 & \underline{0.07}/\underline{0.14}/\underline{0.980} \\
        &   & \textsuperscript{7}0.36/\underline{0.59}/0.907 & 0.39/0.98/0.867 & 0.18/0.49/0.934 & 0.53/0.89/0.864 & 0.43/0.99/0.863 & 0.14/0.29/0.957 \\
  \cmidrule(lr){2-8}      & \multirow{7}[2]{*}{\begin{sideways}PanMamba\end{sideways}} & \textsuperscript{1}2.44/2.33/0.580 & 1.06/1.33/0.779 & 1.54/1.56/0.720 & 2.37/1.31/0.667 & 3.15/1.88/0.565 & 1.96/1.72/0.671 \\
        &   & \textsuperscript{2}0.60/0.91/0.855 & 1.02/1.56/0.759 & 0.49/0.74/0.882 & 1.14/1.39/0.765 & 1.67/2.19/0.656 & 1.03/2.02/0.720 \\
        &   & \textsuperscript{3}0.73/0.97/0.838 & 0.89/1.40/0.785 & 0.51/0.69/0.885 & 1.16/1.34/0.769 & 1.70/2.17/0.658 & 1.16/2.26/0.689 \\
        &   & \textsuperscript{4}0.60/0.91/0.855 & 0.91/1.47/0.776 & 0.44/0.71/0.889 & 1.08/1.35/0.775 & 1.57/2.12/0.670 & 0.95/1.98/0.729 \\
        &   & \textsuperscript{5}0.65/0.92/0.849 & 0.80/1.34/0.798 & 0.50/0.71/0.884 & 1.08/\underline{1.28}/0.782 & 1.60/2.10/0.672 & 1.08/2.22/0.699 \\
        &   & \underline{\textsuperscript{6}0.46}/\underline{0.82}/\underline{0.876} & 0.62/1.29/0.818 & \underline{0.30}/\underline{0.59}/\underline{0.913} & 0.92/1.35/0.788 & 0.98/\underline{1.75}/0.746 & \underline{0.19}/\underline{0.77}/\underline{0.906} \\
        &   & \textsuperscript{7}0.54/0.86/0.865 & \underline{0.53}/\underline{1.20}/\underline{0.834} & 0.37/0.71/0.896 & \underline{0.86}/1.28/\underline{0.800} & \underline{0.87}/1.80/\underline{0.750} & 0.34/1.13/0.857 \\
      \bottomrule
      \end{tabular}}%
    \caption{Quantitative comparison on reduced- and full-resolution evaluation. $\uparrow$: Higher is better, $\downarrow$: Lower is better. \underline{Underlined} values indicate the best results among different configurations of one model. \textbf{Bold} values indicate the best across all models. Index convention: $^1$Training from scratch, $^2$0-shot generalization (pre-trained on ImageNet), $^3$0-shot generalization (pre-trained on SkyScript), $^4$1-shot freeze-tuning (pretrained on ImageNet), $^5$1-shot freeze-tuning (pretrained on SkyScript), $^6$1-shot full-tuning (pretrained on ImageNet), $^7$1-shot full-tuning (pretrained on SkyScript).}
    \label{tab:cmp-reduced-full}%
\end{table*}%

\begin{figure*}[htbp]
  \centering
\vspace{-3mm}
  \includegraphics[width=1.0\linewidth]{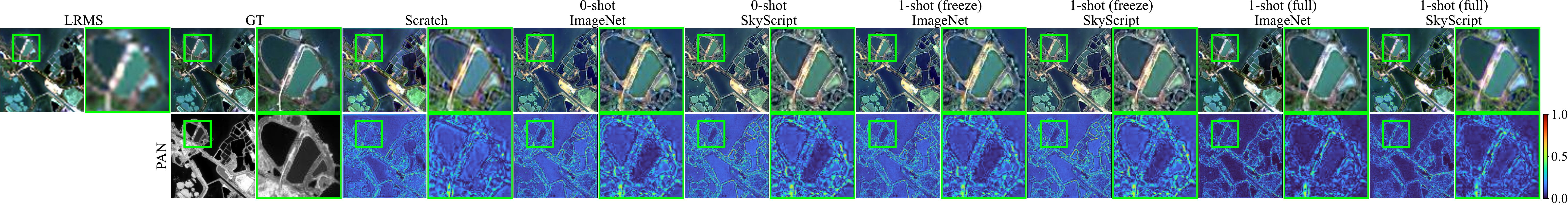} \\
  \includegraphics[width=1.0\linewidth]{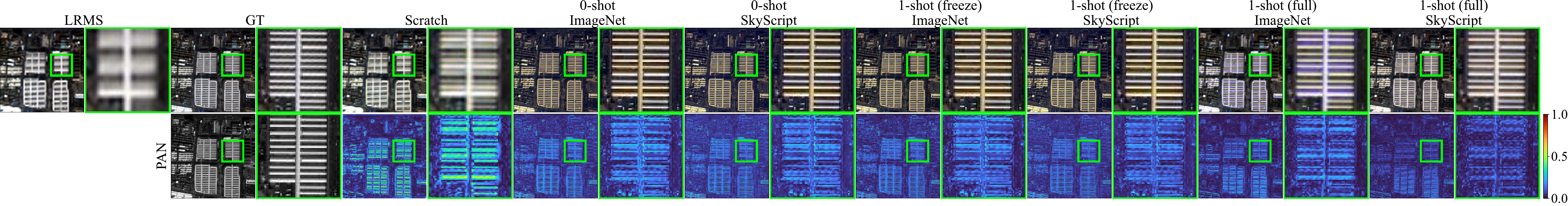}
  \caption{Visualization of predictions and mean absolute errors on IKONOS (upper) and QuickBird (lower) with PEMAE~\cite{PEMAE} model. Bright boxes indicate zoomed-in regions.}
  \label{fig:cmp-viz}
\vspace{-3mm}
\end{figure*}

\subsubsection{Pretraining Results}

Table~\ref{tab:cmp-pretrain-4x8x} summarizes the performance of different pansharpening models pretrained on simulated datasets derived from ImageNet and SkyScript under $4\times$ and $8\times$ spatial resolution degradations. Across all architectures, pretraining on SkyScript consistently yields higher PSNR and SSIM scores compared to ImageNet. Among the models, PEMAE achieves the best performance on both datasets and at both degradation levels, demonstrating the advantage of Transformer-based architectures in capturing long-range dependencies and complex spectral relationships. PanMamba achieves the second-best performance, followed by CNN-based models. Notably, the performance gap between $4\times$ and $8\times$ settings highlights the increased challenge of higher degradation, but the relative ranking of models remains consistent, underscoring the robustness of the pretraining strategy. For visualization of predicted results, please refer to Appendix.

\subsubsection{Reduced-Resolution Evaluation}

Table~\ref{tab:cmp-reduced-full} presents a comprehensive quantitative comparison of different models and training strategies. Several key observations emerge: (1) Models trained from scratch (index $^1$) generally underperform compared to those utilizing pretraining, especially in cross-domain (0-shot) and adaptation (1-shot) scenarios. 0-shot generalization (indices $^2$ and $^3$) already provides a significant boost over training from scratch, demonstrating the transferability of the learned spatial-spectral priors. (2) One-shot full-tuning (indices $^6$ and $^7$) achieves the best results for most models and datasets, indicating that full parameter adaptation enables rapid and effective domain adaptation. (3) Freeze-tuning (indices $^4$ and $^5$) also improves performance but is generally less effective than full-tuning, suggesting that updating all parameters is beneficial for adapting to new domains. This observation differs from high-level vision tasks like classification, where fine-tuning only the classification head often suffices for good performance, highlighting the unique challenges of low-level vision tasks that require pixel-level precision and spatial-spectral alignment. (4) Interestingly, while PEMAE achieves the best performance during pretraining (as shown in Table~\ref{tab:cmp-pretrain-4x8x}), CNN-based models demonstrate superior performance in 0-shot generalization scenarios (indices $^2$ and $^3$). While Transformer-based models excel at learning complex long-range dependencies during pretraining, CNN architectures provide better generalization capabilities for unseen real satellite datasets. This can be attributed to the capabilities in capturing local spatial patterns and inductive bias towards spatial locality, making them more robust when transferring from simulated to real satellite data with varying spectral characteristics.

According to Table~\ref{tab:cmp-complexity}, the performance differences may be related to both total parameter count and the ratio of tunable parameters during fine-tuning. PreMix, with the highest tunable parameter ratio (6.75\%) and largest model size (616.11K parameters), achieves the best performance, suggesting that having more adaptable parameters benefits domain adaptation. Conversely, PEMAE and PanMamba, despite having very low tunable ratios (0.60\% and 0.47\%, respectively), still achieve competitive performance, indicating that the quality of pretrained representations and architectural design can compensate for limited parameter adaptation. These results collectively demonstrate that the proposed pretraining strategy, especially when combined with minimal fine-tuning, substantially enhances the generalization and adaptation capabilities of pansharpening models.

\subsubsection{Full-Resolution Evaluation}

Table~\ref{tab:cmp-reduced-full} also reports the full-resolution evaluation results. The trends observed in reduced-resolution evaluation persist: models with pretraining and subsequent full-tuning (indices $^6$ and $^7$) achieve the best or near-best results across most datasets and models. The results confirm that large-scale simulated pretraining, followed by targeted adaptation, enables models to generalize well to real satellite data and maintain high-quality fusion outputs in challenging full-resolution scenarios.

Notably, GPPNN achieve the best QNR and lowest distortion metrics in most cases, training from scratch. This demonstrates that improving model generalization through meticulous architectural design remains a viable approach, but requires extensive domain-specific engineering. According to Table~\ref{tab:cmp-reduced-full}, our pretraining approach is universally applicable to most models, providing consistent performance improvements across different architectures while requiring minimal real data for adaptation.

\subsubsection{Visualization}

Fig.~\ref{fig:cmp-viz} provides visualization comparison, corroborating the quantitative findings: full-tuning enables the model to rapidly adapt to new spectral mappings while preserving spatial structure, resulting in more accurate and visually pleasing fusion outputs. These visualizations further demonstrate that the proposed pretraining and adaptation strategy is effective in both quantitative and qualitative terms, enabling robust cross-domain pansharpening with minimal real data. Fig.~\ref{fig:cmp-convergence} in Appendix illustrates the convergence behavior. Zero-shot inference (pretrained only) outperforms models trained from scratch, demonstrating strong generalization. Pretraining followed by fine-tuning leads to much faster and higher convergence compared to training from scratch, highlighting the value of learned spatial-spectral priors. For tuning strategy, full-tuning consistently outperforms freeze-tuning, indicating that full adaptation is crucial for optimal performance. Interestingly, pretraining and fine-tuning on ImageNet yields better results than on SkyScript, likely due to the greater diversity and scale of ImageNet, which helps the model learn more generalizable spatial features, while SkyScript provides more domain-specific but potentially less diverse priors.

\section{Conclusion}

In this paper, we proposed pretraining and fine-tuning for pansharpening, leveraging large-scale simulated datasets to learn robust spatial-spectral priors. By pretraining fusion models on natural and remote sensing images and fine-tuning them with extremely limited manner, the models demonstrated significant improvements in generalization performance across different satellite sensors and imaging conditions. Extensive experiments on benchmark datasets showed that the proposed approach enhances the adaptability of models to unseen data with minimal real-world data.

Despite these promising results, while our method improves generalization, it still relies on simulated data, which may not fully capture the complexity of real-world sensor variations. Future research could investigate the incorporation of more diverse datasets, including hyperspectral and synthetic data, to further enhance model robustness. Additionally, exploring more efficient fine-tuning strategies and extending the pretraining paradigm to other remote sensing tasks could lead to more versatile and scalable solutions.

\clearpage
\newpage

\section{Appendix}

\subsection{Datasets}

For pretraining, we utilize two large-scale datasets: ImageNet~\cite{imagenet} and SkyScript~\cite{SkyScript}. For each dataset, we use 40,000 training images and 10,000 validation images. The images are processed to simulate multispectral and panchromatic data pairs for pansharpening pretraining.

For zero-shot and one-shot evaluation, we employ six satellite datasets~\cite{surv2} including WorldView-2, WorldView-3, WorldView-4, IKONOS, QuickBird, and Gaofen-2. Each satellite dataset contains hundreds of image pairs, where we use 10 images for tuning and the remaining images for validation. Training will be conducted following Wald's protocol~\cite{wald}, and testing will be conducted on both reduced- and full-resolution.

\subsection{Metrics}

For reduced-resolution evaluation, we assess model performance using Peak Signal-to-Noise Ratio (PSNR), Spectral Angle Mapper (SAM), Relative Dimensionless Global Error in Synthesis (ERGAS), Correlation Coefficient (CC), Structural Similarity Index Measure (SSIM), and Mean Absolute Error (MAE). For full-resolution evaluation, we employ Spectral Distortion Index ($D_{\lambda}$), Spatial Distortion Index ($D_{S}$), and Quality with No Reference (QNR)~\cite{QNR} to quantitatively measure the quality of the fused images.

\subsection{Implementation Details}

All model implementations are available from the official site. Experiments are conducted using PyTorch $2.1.1$ with $4$ NVIDIA RTX $4090$ GPUs. For pretraining, we employ a batch size of $12$ with AdamW~\cite{loshchilov2018decoupled} optimizer and a learning rate of $10^{-3}$. The training uses linear warmup cosine annealing scheduler for $100$ epochs, with warm up for $10$ epochs. For fine-tuning, we use one single pair of multispectral and panchromatic images with AdamW~\cite{loshchilov2018decoupled} optimizer and a reduced learning rate of $10^{-4}$. The fine-tuning employs cosine annealing scheduler for $40$ epochs. The best model is selected based on the highest PSNR value on validation set.

\subsection{Additional Experimental Results}

\begin{figure}[t]
    \centering
    \includegraphics[width=1.0\linewidth]{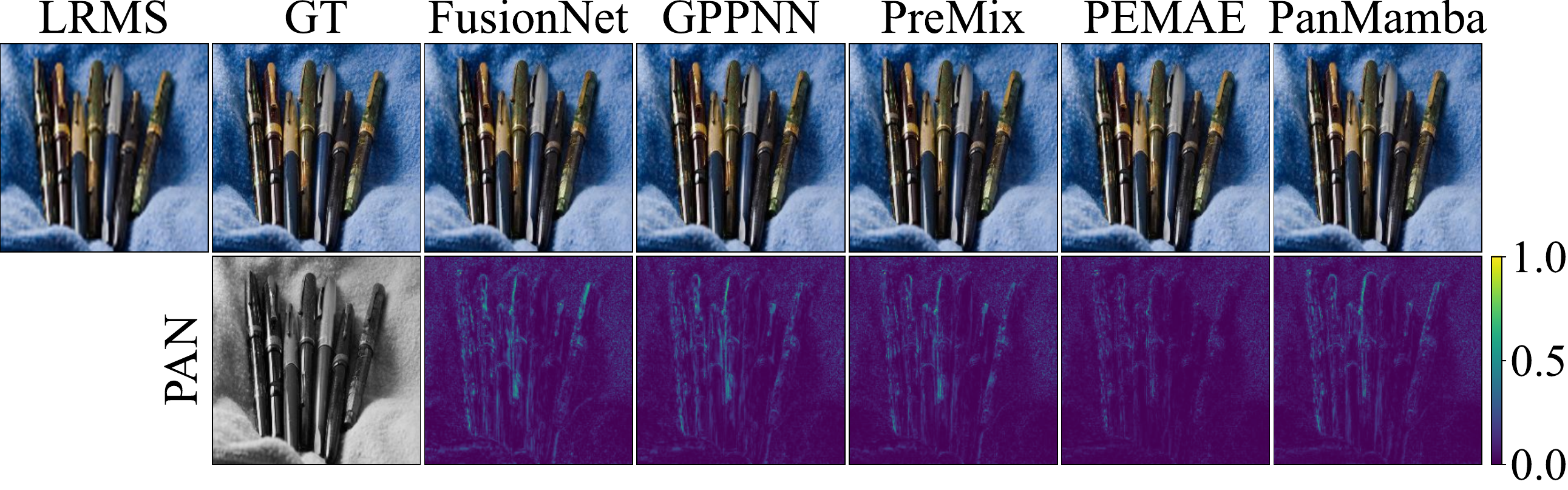} \\
    \includegraphics[width=1.0\linewidth]{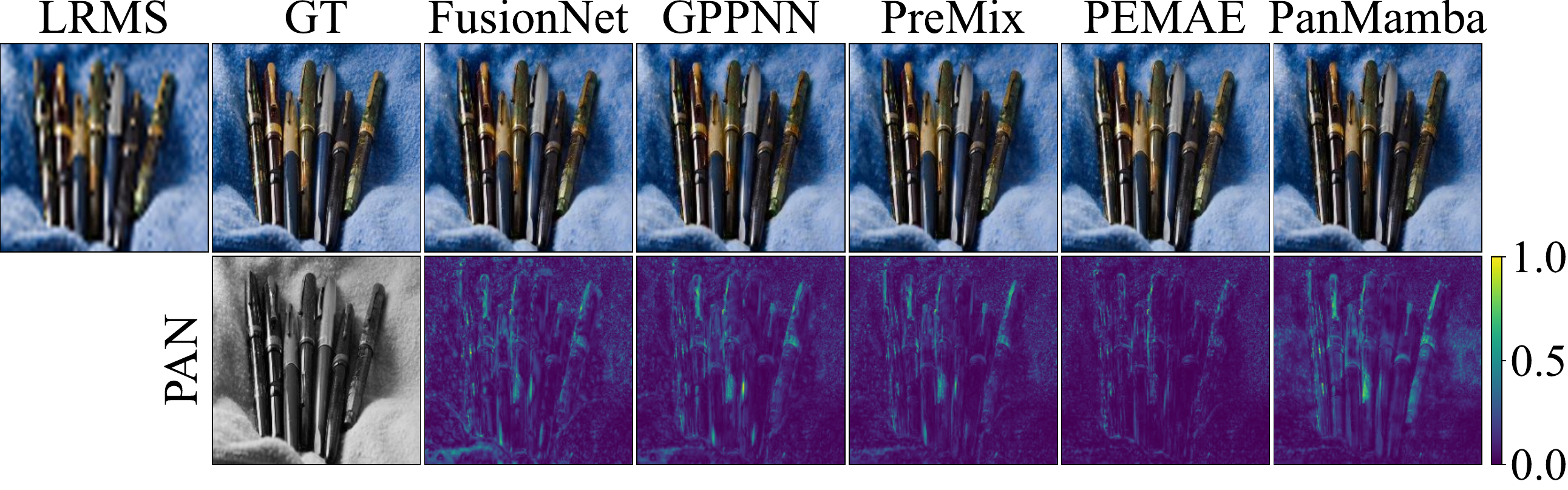}    \caption{Visualization of simulated MS, PAN, predictions, and error maps on ImageNet~\cite{imagenet} with $4\times$ (upper) and $8\times$ (lower) spatial resolution degradations.}
    \label{fig:cmp-pretrain-imagenet-4x8x}
\end{figure}

\begin{figure}[t]
    \centering
    \includegraphics[width=1.0\linewidth]{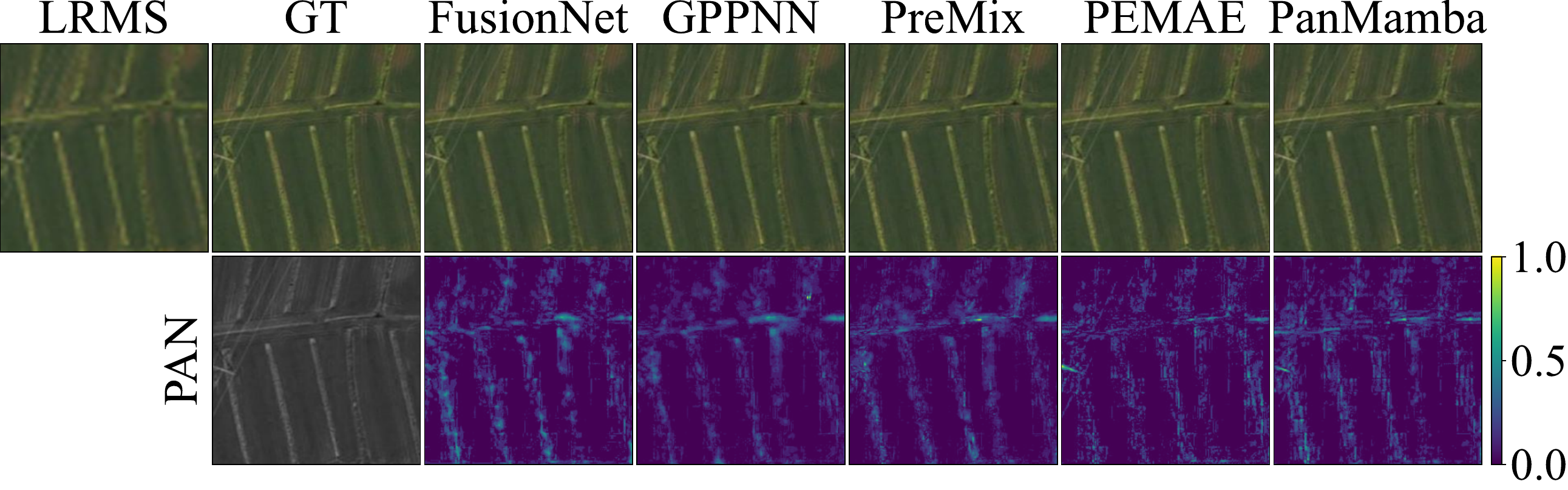} \\
    \includegraphics[width=1.0\linewidth]{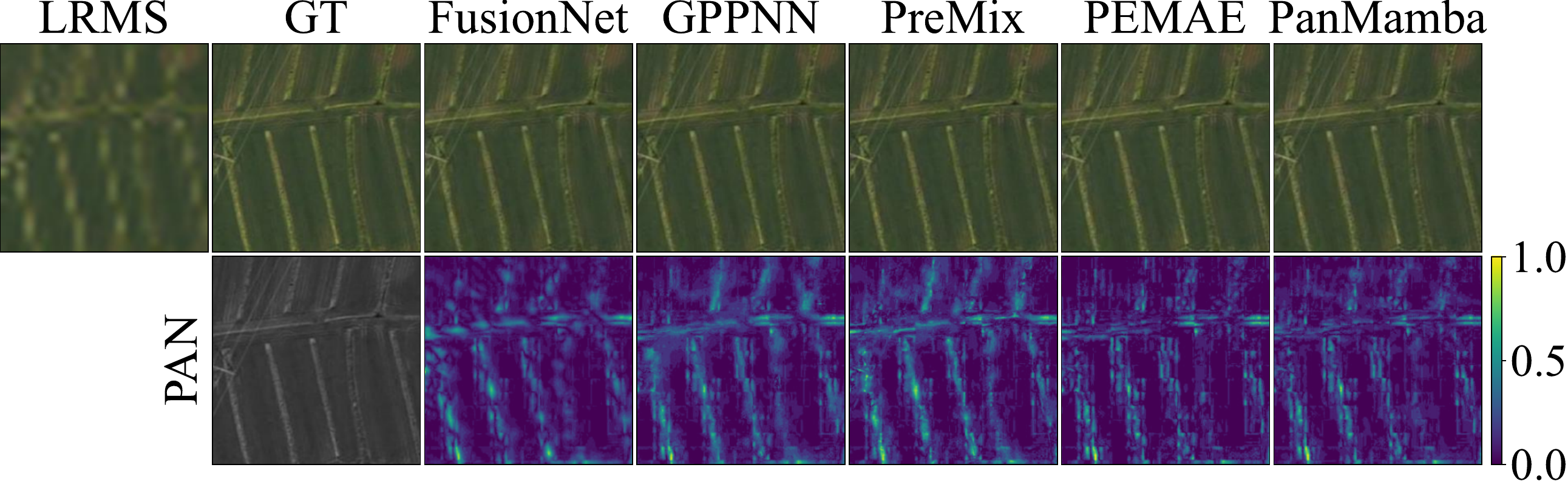}
    \caption{Visualization of simulated MS, PAN, predictions, and error maps on SkyScript~\cite{SkyScript} with $4\times$ (upper) and $8\times$ (lower) spatial resolution degradations.}
    \label{fig:cmp-pretrain-skyscript-4x8x}
\end{figure}

\subsubsection{Quantitative Results.}
Table~\ref{tab:addlabel} presents an additional quantitative comparison (SSIM, MAE, and CC metrics) of different models across six satellite datasets under various training configurations, including training from scratch, zero-shot generalization, and one-shot fine-tuning with both ImageNet and SkyScript pretraining. The results demonstrate that our proposed method achieves superior performance in most scenarios, particularly in the one-shot learning settings.

\begin{table*}[t]
    \centering
    \resizebox{0.9\textwidth}{!}{
      \begin{tabular}{ccllllll}
      \toprule
      \multicolumn{2}{c}{Model} & \multicolumn{1}{c}{WorldView-2} & \multicolumn{1}{c}{WorldView-3} & \multicolumn{1}{c}{WorldView-4} & \multicolumn{1}{c}{IKONOS} & \multicolumn{1}{c}{Quickbird} & \multicolumn{1}{c}{Gaofen-2} \\
      \midrule
      \multirow{35}[10]{*}{\begin{sideways}Reduced-Resolution : SS$\uparrow$ / MAE$^{\times 10^{-2}}\downarrow$ / CC$\uparrow$\end{sideways}} & \multirow{7}[2]{*}{\begin{sideways}FusionNet\end{sideways}} & \textsuperscript{1}0.793/2.38/0.824 & 0.859/1.98/0.892 & 0.913/1.26/0.925 & 0.888/1.23/0.904 & 0.958/0.63/0.907 & 0.928/0.87/0.939 \bigstrut[t]\\
      &   & \textsuperscript{2}0.847/1.82/0.877 & 0.827/2.24/0.857 & 0.891/1.34/0.905 & 0.879/1.24/0.878 & 0.909/0.82/0.825 & 0.786/1.60/0.788 \\
      &   & \textsuperscript{3}0.846/1.85/0.875 & 0.825/2.20/0.861 & 0.889/1.40/0.902 & 0.879/1.25/0.878 & 0.910/0.83/0.824 & 0.779/1.67/0.778 \\
      &   & \textsuperscript{4}0.857/1.76/0.885 & 0.834/2.17/0.864 & 0.901/1.26/0.914 & 0.888/1.18/0.886 & 0.922/0.75/0.841 & 0.812/1.43/0.819 \\
      &   & \textsuperscript{5}0.852/1.80/0.880 & 0.833/2.13/0.867 & 0.896/1.34/0.908 & 0.886/1.20/0.885 & 0.921/0.77/0.837 & 0.800/1.54/0.800 \\
      &   & \underline{\textsuperscript{6}0.886}/\underline{1.60}/0.905 & \underline{0.866}/\underline{1.93}/0.898 & 0.919/1.11/0.933 & \underline{0.913}/\underline{1.00}/\underline{0.919} & 0.963/0.49/\underline{0.924} & \underline{0.930}/\underline{0.78}/\underline{0.945} \\
      &   & \textsuperscript{7}0.884/1.61/\underline{0.906} & 0.866/1.94/\underline{0.902} & \underline{0.922}/\underline{1.11}/\underline{0.941} & 0.911/1.01/0.919 & \underline{0.964}/\underline{0.48}/0.924 & 0.925/0.81/0.943 \bigstrut[b]\\
\cline{2-8}      & \multirow{7}[2]{*}{\begin{sideways}GPPNN\end{sideways}} & \textsuperscript{1}0.817/2.18/0.862 & 0.833/2.14/0.885 & 0.896/1.26/0.936 & 0.849/1.29/0.888 & 0.945/0.55/0.911 & \underline{0.929}/\underline{0.77}/\underline{0.946} \bigstrut[t]\\
      &   & \textsuperscript{2}0.870/1.73/0.894 & 0.807/2.40/0.851 & 0.885/1.40/0.894 & 0.880/1.23/0.878 & 0.910/0.80/0.828 & 0.781/1.62/0.785 \\
      &   & \textsuperscript{3}0.864/1.80/0.889 & 0.810/2.37/0.855 & 0.875/1.60/0.886 & 0.879/1.23/0.878 & 0.910/0.81/0.832 & 0.779/1.66/0.786 \\
      &   & \textsuperscript{4}0.879/1.64/0.902 & 0.834/2.13/0.869 & 0.905/1.21/0.918 & 0.893/1.13/0.896 & 0.938/0.63/0.867 & 0.850/1.21/0.857 \\
      &   & \textsuperscript{5}0.879/1.67/0.900 & 0.843/2.09/0.877 & 0.903/1.29/0.917 & 0.902/1.07/0.906 & 0.944/0.59/0.887 & 0.847/1.23/0.859 \\
      &   & \textsuperscript{6}0.885/1.58/0.907 & 0.847/2.01/0.879 & 0.919/\underline{1.09}/\underline{0.941} & 0.906/1.03/\underline{0.914} & 0.951/0.54/0.908 & 0.927/0.79/0.941 \\
      &   & \underline{\textsuperscript{7}0.887}/\underline{1.57}/\underline{0.909} & \underline{0.857}/\underline{1.96}/\underline{0.894} & \underline{0.920}/1.09/0.939 & \underline{0.908}/\underline{1.03}/0.912 & \underline{0.962}/\underline{0.49}/\underline{0.927} & 0.928/0.78/0.945 \bigstrut[b]\\
\cline{2-8}      & \multirow{7}[2]{*}{\begin{sideways}PreMix\end{sideways}} & \textsuperscript{1}0.762/2.25/0.803 & 0.818/2.46/0.875 & 0.885/1.39/0.921 & 0.853/1.37/0.884 & 0.944/0.59/0.896 & 0.930/0.79/0.944 \bigstrut[t]\\
      &   & \textsuperscript{2}0.886/1.59/0.905 & 0.848/2.07/0.873 & 0.908/1.22/0.920 & 0.895/1.14/0.895 & 0.921/0.75/0.843 & 0.815/1.49/0.808 \\
      &   & \textsuperscript{3}0.873/1.76/0.892 & 0.837/2.21/0.867 & 0.898/1.35/0.907 & 0.885/1.21/0.883 & 0.914/0.79/0.830 & 0.800/1.57/0.795 \\
      &   & \textsuperscript{4}0.894/1.54/0.909 & 0.859/1.99/0.882 & 0.914/1.16/0.926 & 0.904/1.07/0.903 & 0.935/0.67/0.861 & 0.844/1.32/0.845 \\
      &   & \textsuperscript{5}0.881/1.70/0.898 & 0.850/2.09/0.877 & 0.906/1.26/0.917 & 0.896/1.14/0.893 & 0.929/0.71/0.848 & 0.827/1.42/0.824 \\
      &   & \underline{\textsuperscript{6}0.901}/\underline{1.46}/\underline{\textbf{0.919}} & \underline{0.877}/\underline{1.83}/\underline{\textbf{0.906}} & \underline{\textbf{0.925}}/\underline{\textbf{1.05}}/\underline{0.942} & \underline{0.918}/\underline{0.97}/\underline{0.923} & \underline{0.965}/\underline{0.47}/\underline{0.931} & \underline{\textbf{0.935}}/\underline{\textbf{0.75}}/\underline{\textbf{0.947}} \\
      &   & \textsuperscript{7}0.892/1.55/0.911 & 0.867/1.91/0.898 & 0.923/1.08/0.939 & 0.909/1.04/0.912 & 0.962/0.49/0.928 & 0.929/0.78/0.945 \bigstrut[b]\\
\cline{2-8}      & \multirow{7}[2]{*}{\begin{sideways}PEMAE\end{sideways}} & \textsuperscript{1}0.781/2.48/0.739 & 0.787/2.58/0.805 & 0.847/1.83/0.816 & 0.871/1.30/0.866 & 0.955/0.58/0.878 & 0.917/0.93/0.904 \bigstrut[t]\\
      &   & \textsuperscript{2}0.882/1.60/0.882 & 0.841/2.06/0.858 & 0.901/1.25/0.908 & 0.893/1.14/0.892 & 0.925/0.73/0.826 & 0.809/1.49/0.802 \\
      &   & \textsuperscript{3}0.875/1.70/0.890 & 0.830/2.15/0.863 & 0.892/1.36/0.908 & 0.885/1.17/0.887 & 0.917/0.75/0.833 & 0.797/1.53/0.796 \\
      &   & \textsuperscript{4}0.886/1.57/0.886 & 0.849/2.01/0.865 & 0.907/1.20/0.914 & 0.899/1.10/0.898 & 0.932/0.69/0.837 & 0.824/1.40/0.820 \\
      &   & \textsuperscript{5}0.879/1.68/0.894 & 0.842/2.07/0.873 & 0.900/1.30/0.916 & 0.894/1.12/0.896 & 0.928/0.69/0.849 & 0.817/1.42/0.818 \\
      &   & \underline{\textbf{\textsuperscript{6}0.903}}/\underline{\textbf{1.46}}/\underline{0.904} & \underline{\textbf{0.878}}/\underline{\textbf{1.82}}/\underline{0.900} & 0.922/\underline{1.07}/\underline{\textbf{0.944}} & \underline{\textbf{0.922}}/\underline{\textbf{0.94}}/\underline{\textbf{0.934}} & \underline{\textbf{0.968}}/\underline{\textbf{0.46}}/\underline{\textbf{0.934}} & \underline{0.930}/\underline{0.77}/0.946 \\
      &   & \textsuperscript{7}0.894/1.55/0.900 & 0.865/1.90/0.897 & \underline{0.924}/1.08/0.941 & 0.918/0.96/0.928 & 0.967/0.46/0.932 & 0.929/0.77/\underline{0.946} \bigstrut[b]\\
\cline{2-8}      & \multirow{7}[2]{*}{\begin{sideways}PanMamba\end{sideways}} & \textsuperscript{1}0.590/6.45/0.549 & 0.698/4.94/0.809 & 0.752/7.04/0.837 & 0.749/5.88/0.794 & 0.884/3.17/0.584 & 0.809/6.58/0.762 \bigstrut[t]\\
      &   & \textsuperscript{2}0.863/1.77/0.885 & 0.819/2.27/0.851 & 0.886/1.41/0.894 & 0.876/1.27/0.877 & 0.916/0.81/0.820 & 0.779/1.66/0.774 \\
      &   & \textsuperscript{3}0.845/1.97/0.882 & 0.812/2.35/0.856 & 0.878/1.57/0.891 & 0.870/1.31/0.868 & 0.907/0.84/0.814 & 0.769/1.72/0.765 \\
      &   & \textsuperscript{4}0.863/1.77/0.885 & 0.823/2.24/0.853 & 0.888/1.40/0.894 & 0.879/1.25/0.878 & 0.920/0.78/0.822 & 0.786/1.63/0.777 \\
      &   & \textsuperscript{5}0.851/1.95/0.883 & 0.820/2.29/0.858 & 0.880/1.56/0.891 & 0.873/1.29/0.871 & 0.913/0.81/0.818 & 0.778/1.67/0.770 \\
      &   & \underline{\textsuperscript{6}0.885}/\underline{1.62}/0.896 & \underline{0.861}/\underline{1.96}/0.883 & \underline{0.914}/\underline{1.21}/\underline{0.921} & 0.895/\underline{1.16}/\underline{0.897} & 0.957/0.58/0.900 & \underline{0.913}/\underline{0.98}/\underline{0.921} \\
      &   & \textsuperscript{7}0.879/1.71/\underline{0.899} & 0.852/2.04/\underline{0.883} & 0.902/1.40/0.909 & \underline{0.895}/1.16/0.896 & \underline{0.957}/\underline{0.56}/\underline{0.906} & 0.904/1.03/0.911 \bigstrut[b]\\
    \bottomrule
      \end{tabular}}
    \caption{Quantitative comparison on reduced-resolution evaluation with metrics: SSIM$\uparrow$ / MAE$^{\times 10^{-2}}\downarrow$ / CC$\uparrow$. $\uparrow$: Higher is better, $\downarrow$: Lower is better. \underline{Underlined} values indicate the best results among different configurations of one model. \textbf{Bold} values indicate the best across all models. Index convention: $^1$Training from scratch, $^2$0-shot generalization (pre-trained on ImageNet), $^3$0-shot generalization (pre-trained on SkyScript), $^4$1-shot freeze-tuning (pretrained on ImageNet), $^5$1-shot freeze-tuning (pretrained on SkyScript), $^6$1-shot full-tuning (pretrained on ImageNet), $^7$1-shot full-tuning (pretrained on SkyScript).}
    \label{tab:addlabel}%
  \end{table*}%
  
  \begin{figure*}[b]
      \centering
      \includegraphics[width=1\linewidth]{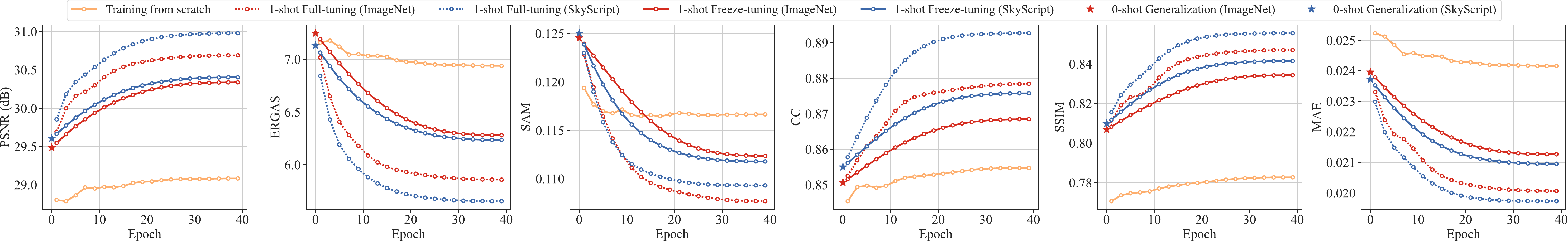} \\
      \caption{Convergence of GPPNN~\cite{GPPNN} on WorldView-3 (averaged results on all training images).}
      \label{fig:cmp-convergence}
      \end{figure*}

\subsubsection{Pretraining Visualization Results.}

Figs.~\ref{fig:cmp-pretrain-imagenet-4x8x} and~\ref{fig:cmp-pretrain-skyscript-4x8x} demonstrate the effectiveness of our pretraining strategy on ImageNet and SkyScript datasets, respectively. These visualizations show the simulated multispectral (MS) and panchromatic (PAN) images, along with the model predictions and corresponding error maps.

\begin{figure*}[t]
    \centering
    \includegraphics[width=1.0\linewidth]{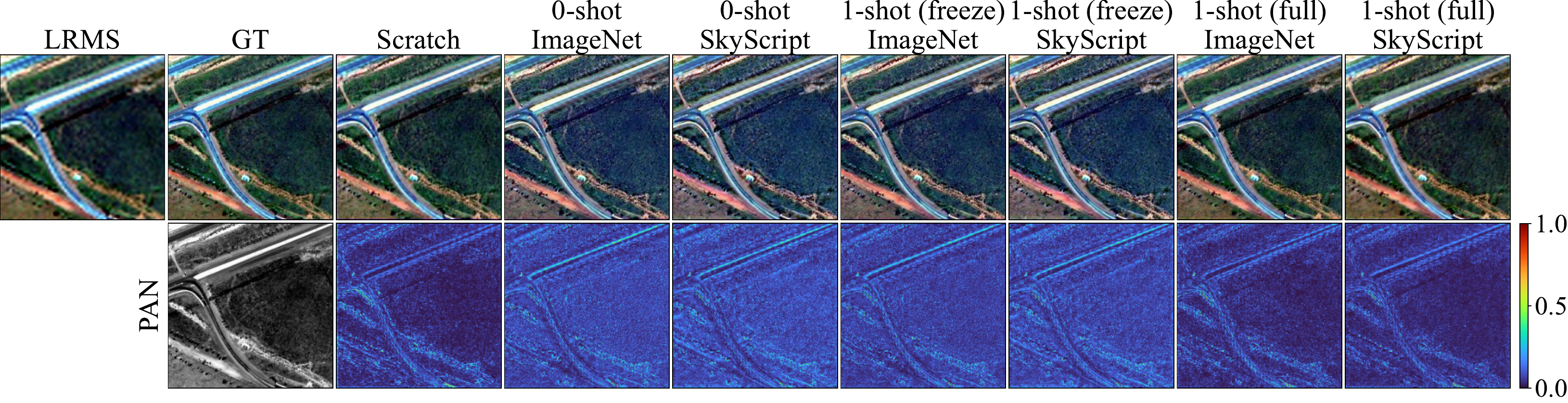} \\
    \vspace{-0.2cm}
    \text{(a) FusionNet} \\
    \vspace{0.1cm}
    \includegraphics[width=1.0\linewidth]{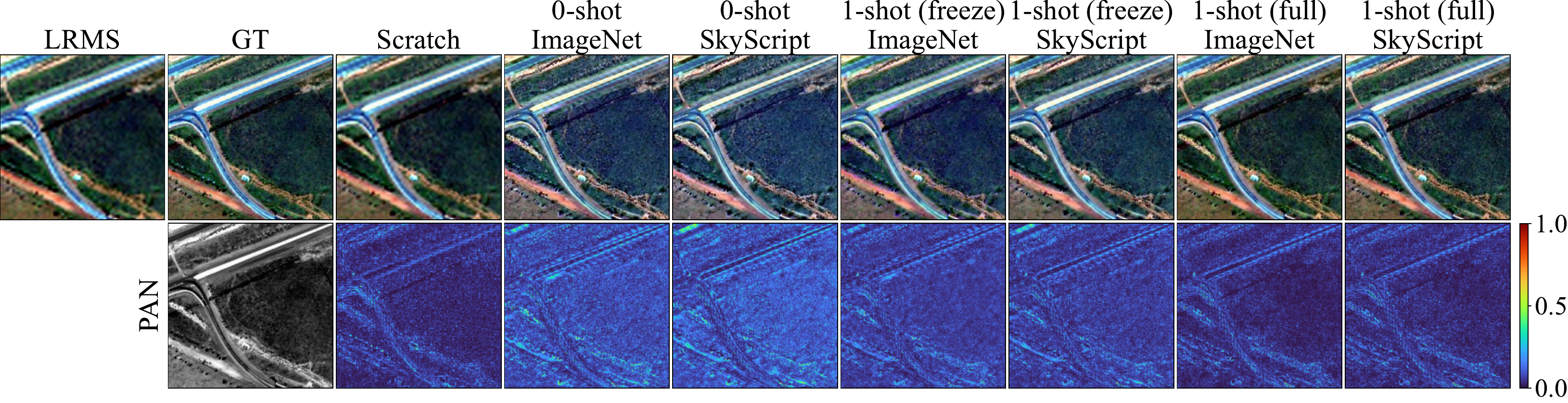} \\
    \vspace{-0.2cm}
    \text{(b) GPPNN} \\
    \vspace{0.1cm}
    \includegraphics[width=1.0\linewidth]{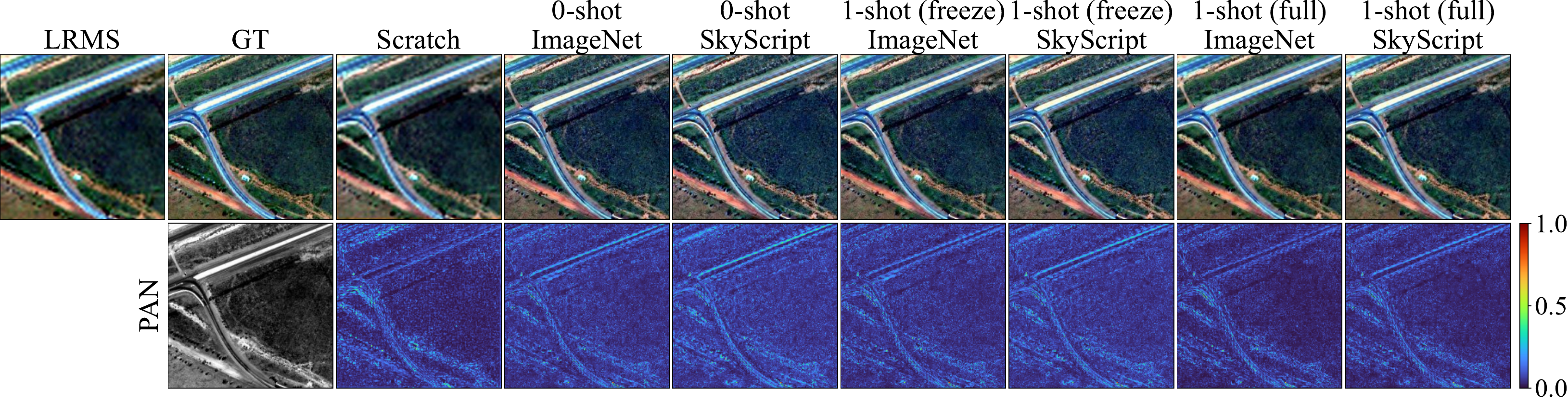} \\
    \vspace{-0.2cm}
    \text{(c) PreMix} \\
    \vspace{0.1cm}
    \includegraphics[width=1.0\linewidth]{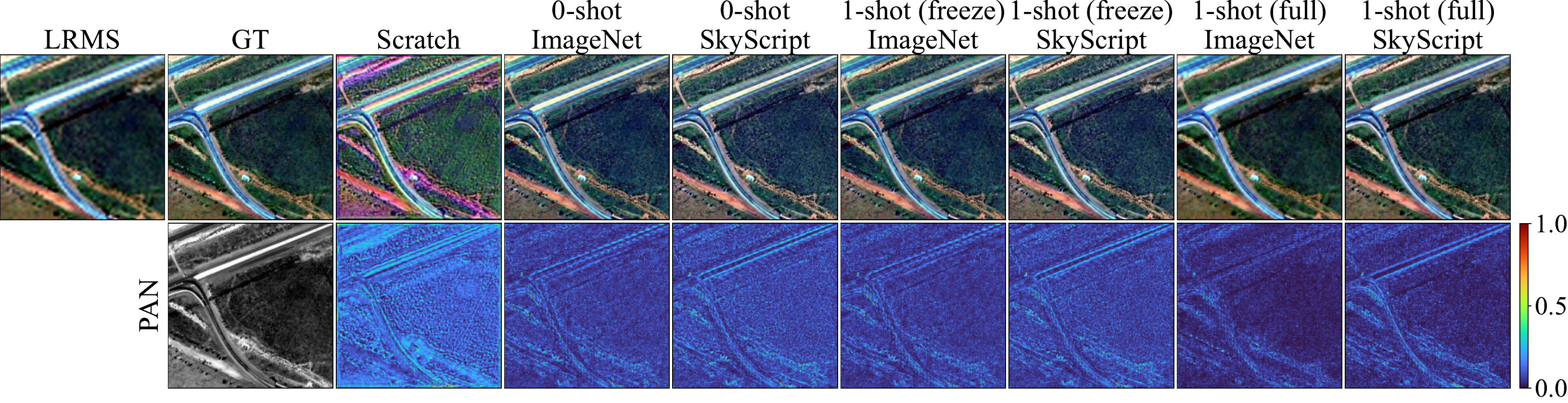} \\
    \vspace{-0.2cm}
    \text{(d) PEMAE} \\
    \vspace{0.1cm}
    \includegraphics[width=1.0\linewidth]{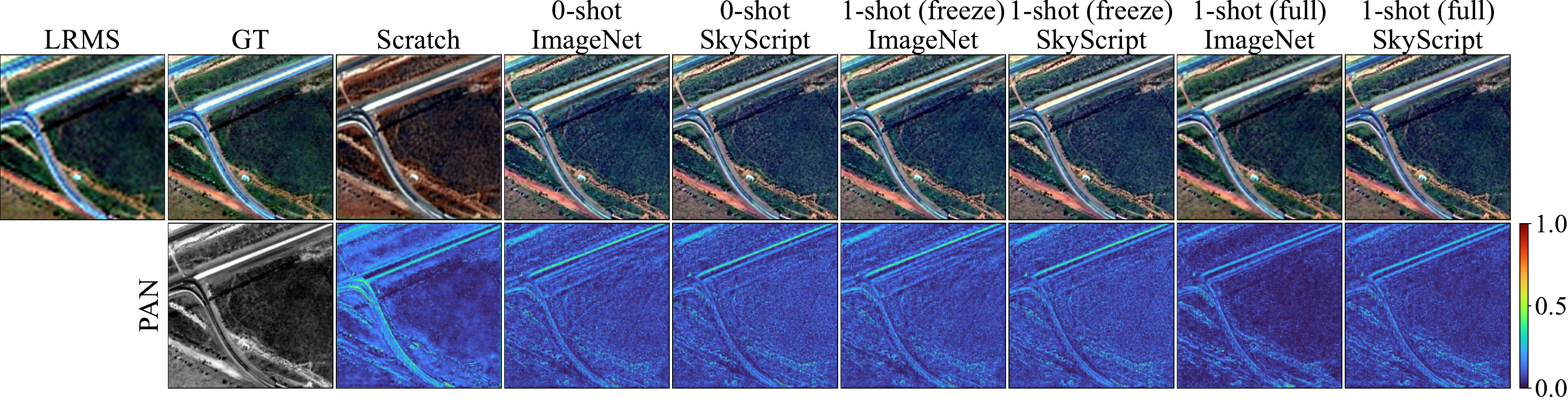} \\
    \vspace{-0.2cm}
    \text{(e) PanMamba} \\
    \vspace{-0.1cm}
    \caption{Visual comparison of different models on WorldView-4.}
    \label{fig:cmp-reduce-wv4}
\end{figure*}

\begin{figure*}[t]
    \centering
    \includegraphics[width=1.0\linewidth]{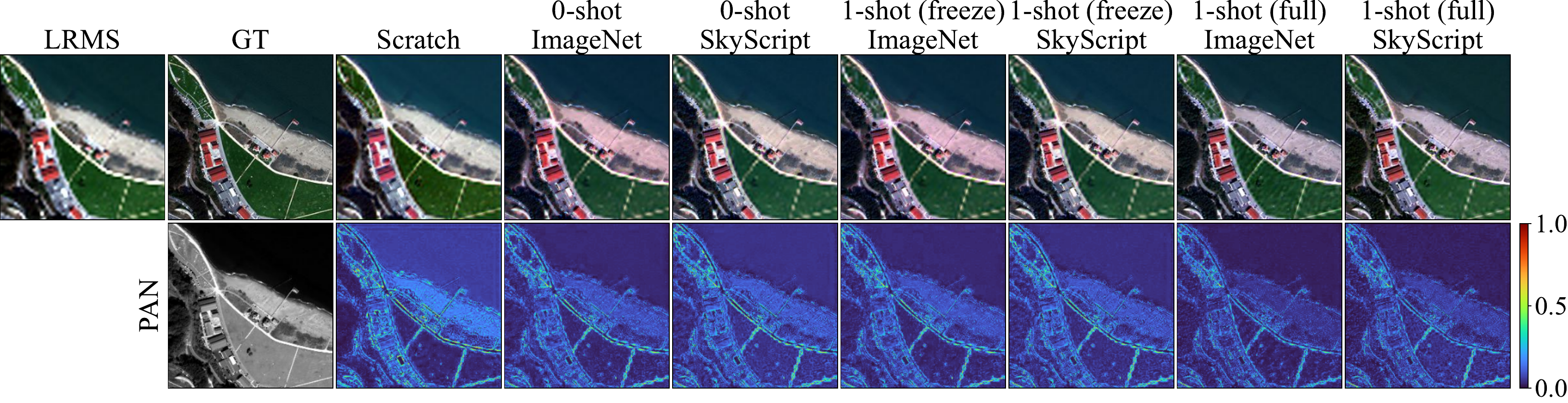} \\
    \vspace{-0.2cm}
    \text{(a) FusionNet} \\
    \vspace{0.1cm}
    \includegraphics[width=1.0\linewidth]{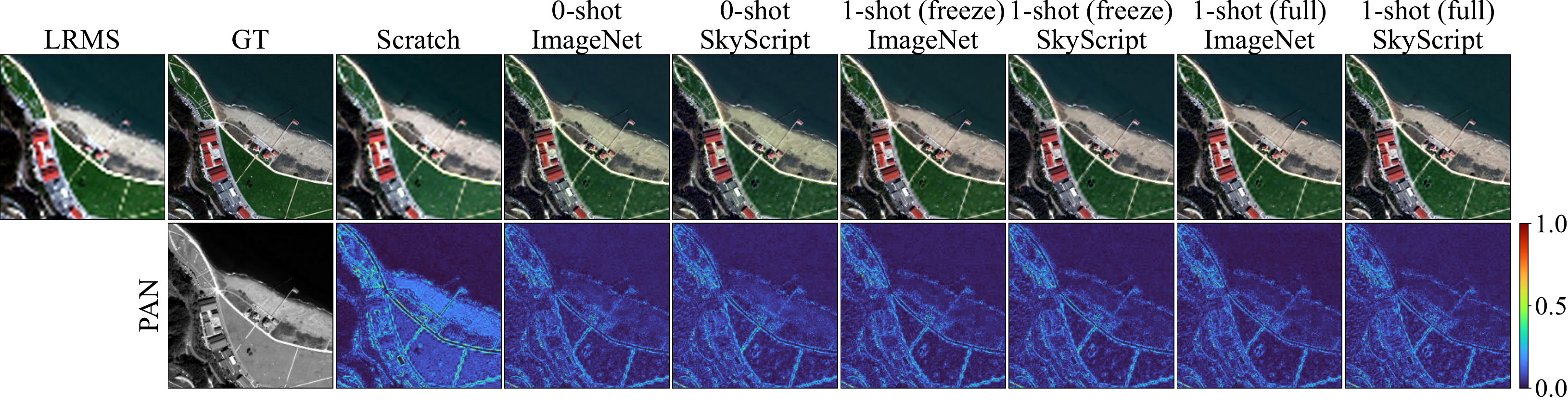} \\
    \vspace{-0.2cm}
    \text{(b) GPPNN} \\
    \vspace{0.1cm}
    \includegraphics[width=1.0\linewidth]{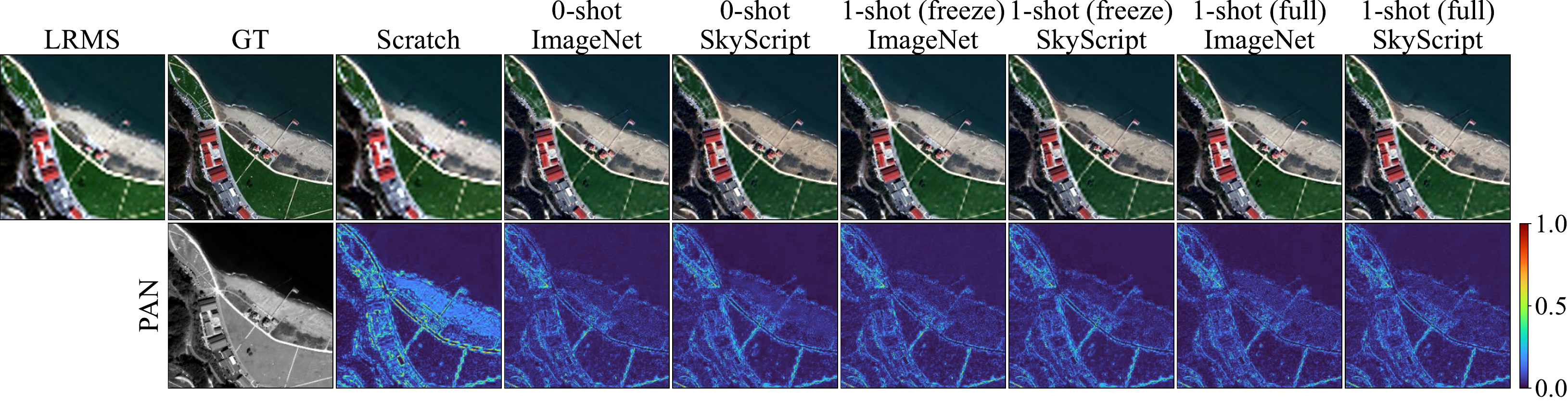} \\
    \vspace{-0.2cm}
    \text{(c) PreMix} \\
    \vspace{0.1cm}
    \includegraphics[width=1.0\linewidth]{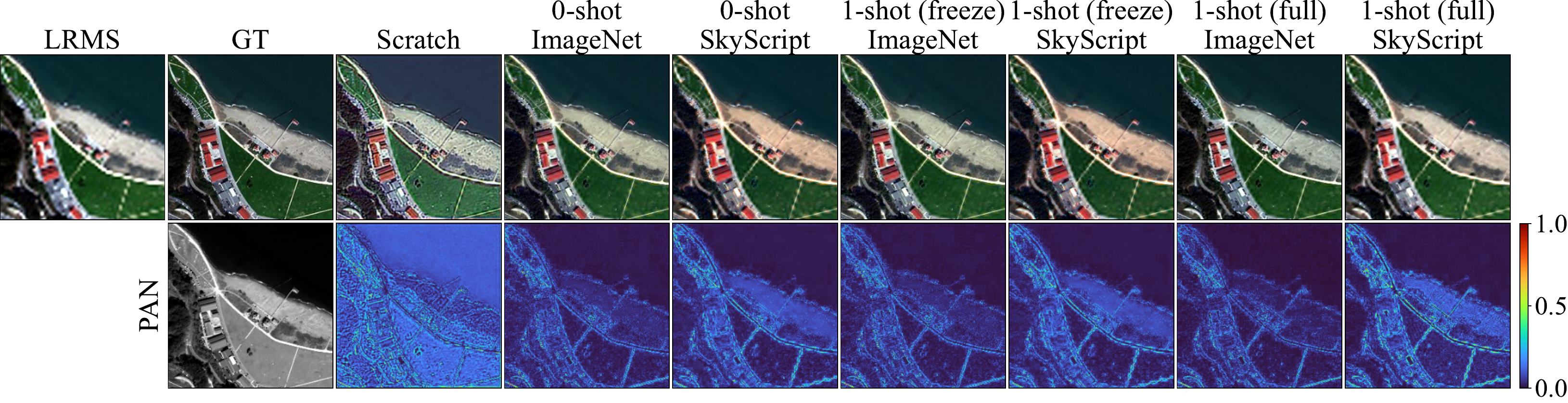} \\
    \vspace{-0.2cm}
    \text{(d) PEMAE} \\
    \vspace{0.1cm}
    \includegraphics[width=1.0\linewidth]{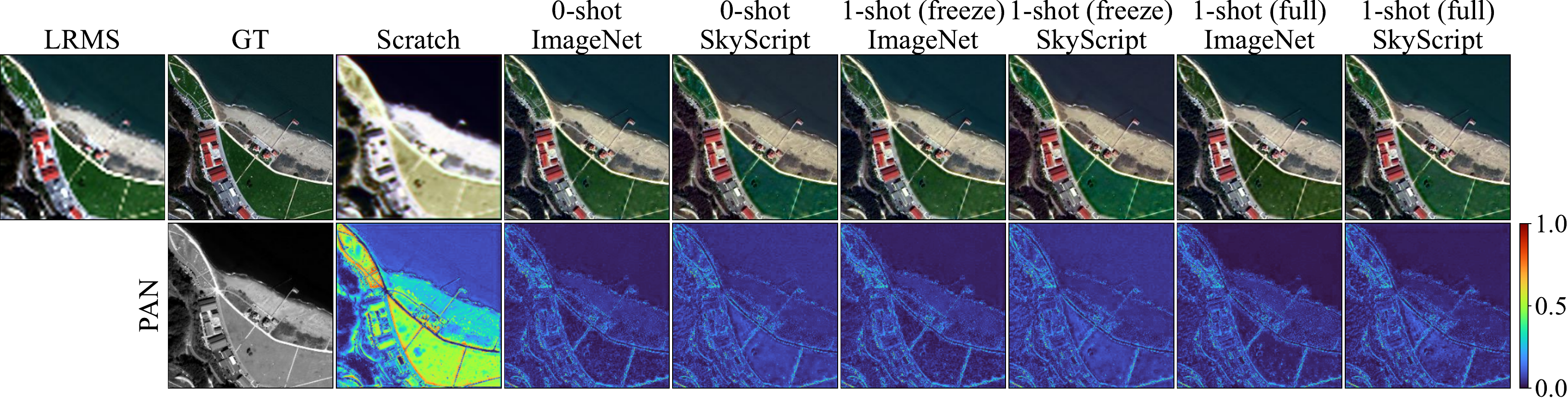} \\
    \vspace{-0.2cm}
    \text{(e) PanMamba} \\
    \vspace{-0.1cm}
    \caption{Visual comparison of different models on WorldView-2.}
    \label{fig:cmp-reduce-wv2}
\end{figure*}

Figs.~\ref{fig:cmp-reduce-wv4} and~\ref{fig:cmp-reduce-wv2} provide additional visual comparisons of different pansharpening methods on WorldView-4 and WorldView-2 datasets, respectively. The visual results clearly demonstrate the superior performance of pretraining and fine-tuning in preserving both spectral information and spatial details. Our approach shows better edge preservation, reduced artifacts, and more accurate spectral reproduction.

\clearpage

\bibliography{aaai2026}

\end{document}